\documentclass[11pt, a4paper,  copyright, nonumbering]{techreport}
\usepackage[authoryear, sort&compress, round]{natbib}
\usepackage{dblfloatfix}
\usepackage{ulem}
\usepackage{caption}
\usepackage{dramatist}
\usepackage{xspace}
\usepackage{pifont} %
\usepackage{multirow}
\usepackage{tcolorbox}
\usepackage{xltabular}
\usepackage{longtable}
\usepackage{hyperref}
\usepackage{subfig}

\usepackage{wrapfig} 

\usepackage{amsfonts}
\usepackage{amsmath}
\usepackage{amssymb}
\usepackage{lineno}
\usepackage{multirow}
\usepackage{adjustbox}

\usepackage[bottom]{footmisc}

\usepackage{CJKutf8}
\usepackage{setspace}

\usepackage{dsfont}
\usepackage{array} %
\usepackage{tabularx} %
\usepackage{xcolor} %
\usepackage{tabularx}
\usepackage{booktabs}

\usepackage{lipsum}  %
\usepackage{multicol} %

\usepackage{microtype}
\usepackage{subcaption}  
\usepackage{graphicx}
\usepackage{algpseudocode}
\usepackage{bbm}
\usepackage{algorithm}
\usepackage{url}
\usepackage{caption}
\usepackage{tabu}
\usepackage{adjustbox}
\usepackage{rotating}
\usepackage{arydshln}
\usepackage{wrapfig}
\usepackage{pgf}
\usepackage{float}
\usepackage{flafter}
\usepackage{enumitem}
\usepackage{ifthen}

\newboolean{arxiv}
\setboolean{arxiv}{true}
\newboolean{icmlfinal}
\setboolean{icmlfinal}{false}

\usepackage{booktabs} 

\usepackage{amsmath}
\usepackage{amssymb}
\usepackage{mathtools}
\usepackage{amsthm}

\usepackage[capitalize,noabbrev]{cleveref}

\usepackage[final]{changes}

\crefname{equation}{Eq.}{Eqs.}
\Crefname{equation}{Eq.}{Eqs.}
\crefname{figure}{Fig.}{Figs.}
\Crefname{figure}{Fig.}{Figs.}

\usepackage{amsmath,amsfonts,bm}

\def\eqref#1{equation~\ref{#1}}

\def\1{\bm{1}}

\DeclareMathAlphabet{\mathsfit}{\encodingdefault}{\sfdefault}{m}{sl}
\SetMathAlphabet{\mathsfit}{bold}{\encodingdefault}{\sfdefault}{bx}{n}

\theoremstyle{plain}

\theoremstyle{definition}

\theoremstyle{remark}

\usepackage{todonotes}

\makeatletter
\def\@BTrule[#1]{%
  \ifx\longtable\undefined
    \let\@BTswitch\@BTnormal
  \else\ifx\hline\LT@hline
    \nobreak
    \let\@BTswitch\@BLTrule
  \else
     \let\@BTswitch\@BTnormal
  \fi\fi
  \global\@thisrulewidth=#1\relax
  \ifnum\@thisruleclass=\tw@\vskip\@aboverulesep\else
  \ifnum\@lastruleclass=\z@\vskip\@aboverulesep\else
  \ifnum\@lastruleclass=\@ne\vskip\doublerulesep\fi\fi\fi
  \@BTswitch}
\makeatother

\addto\extrasenglish{
}

 {\begin{list}{}%
         {\setlength{\leftmargin}{#1}}%
         \item[]%
 }
 {\end{list}}

\renewcommand{\today}{}
\title{\centering Unified Multimodal Discrete Diffusion}

\title{\centering Unified Multimodal Discrete Diffusion}

\author{\Authfont Alexander Swerdlow* \qquad Mihir Prabhudesai* \qquad Siddharth Gandhi \\ 
\Authfont Deepak Pathak \qquad Katerina Fragkiadaki\\  \vspace{10pt}
\Affilfont Carnegie Mellon University}

\footnotetext[1]{*Equal contribution}
\footnotetext[2]{Correspondence to: aswerdlow@cmu.edu, mprabhud@andrew.cmu.edu}

\newcommand{\model}{\text{UniDisc}}

\begin{abstract}
    Multimodal generative models that can understand and generate across multiple modalities are dominated by autoregressive (AR) approaches, which process tokens sequentially from left to right, or top to bottom. These models jointly handle images, text, video, and audio for various tasks such as image captioning, question answering, and image generation. In this work, we explore discrete diffusion models as a unified generative formulation in the joint text and image domain, building upon their recent success in text generation. Discrete diffusion models offer several advantages over AR models, including improved control over quality versus diversity of generated samples, the ability to perform joint multimodal inpainting (across both text and image domains), and greater controllability in generation through guidance. Leveraging these benefits, we present the first \textbf{Uni}fied Multimodal \textbf{Disc}rete Diffusion (\model{}) model which is capable of jointly understanding and generating text and images for a variety of downstream tasks. We compare \model{} to multimodal AR models, performing a scaling analysis and demonstrating that \model{} outperforms them in terms of both performance and inference-time compute, enhanced controllability, editability, inpainting, and flexible trade-off between inference time and generation quality. Code and additional visualizations are available at \url{https://unidisc.github.io}.
\end{abstract}

\begin{document}
\setboolean{arxiv}{true} 

\maketitle

\section{Introduction}
\label{intro}
Multimodal generative models—which can understand and generate a variety of modalities such as text, images, videos, and audio – can significantly improve the capabilities of an AI system, as these models can (1) leverage information from multiple sources to better understand the context (2) learn from any available data source, and (3) respond to a user's request in a flexible manner, thus dynamically generating text, images, or audio as required. Although the choice of model architecture—transformers—is currently clear, the optimal generative objective remains unclear.

\begin{figure}[h]
    \centering
    \includegraphics[width=1.0\linewidth]{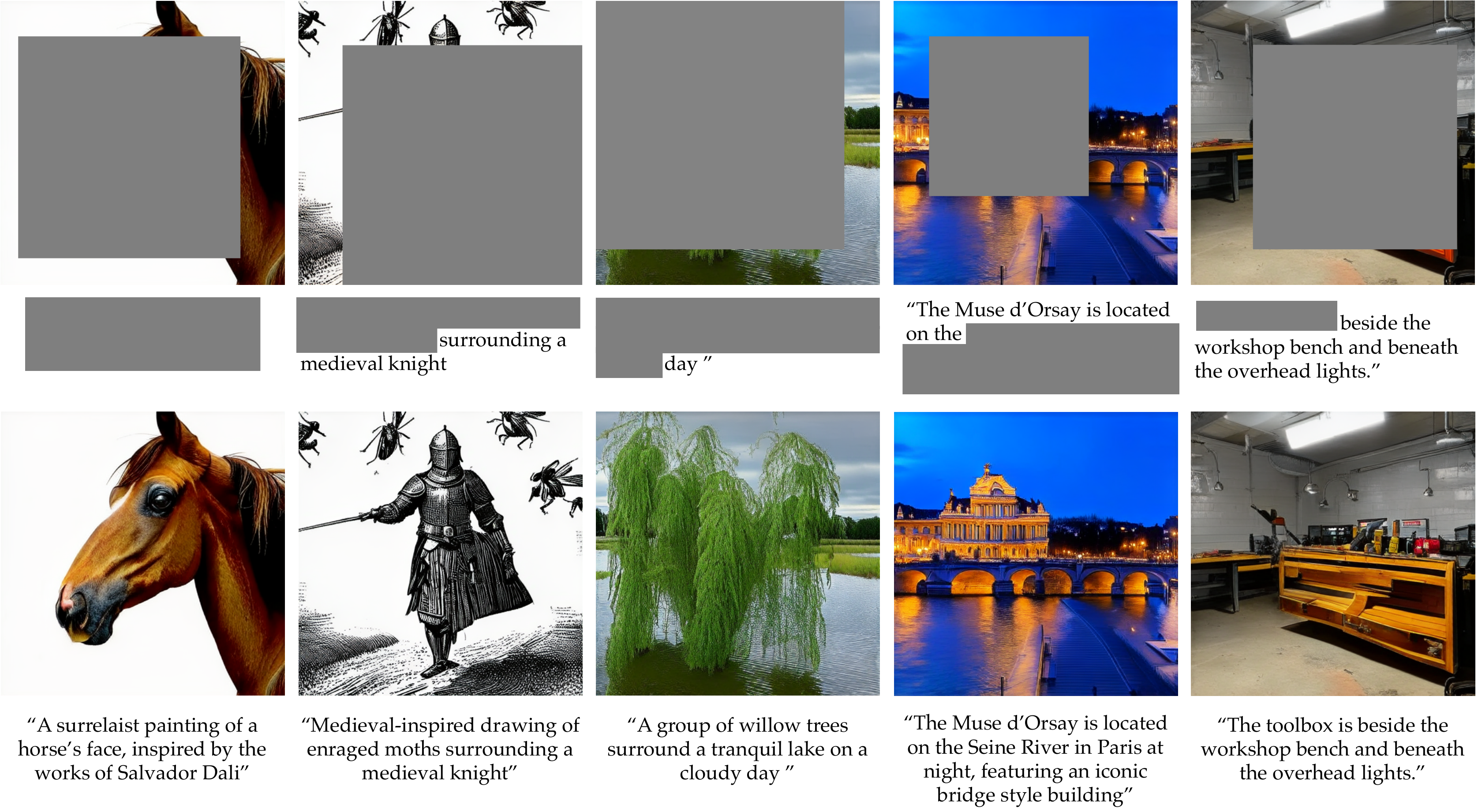}
    \vspace{-17pt}
    \caption{\textbf{We show \model{}'s ability to jointly inpaint image \& text pairs}. We do not explicitly optimize for this objective but it is intrinsic to \model{}'s unified diffusion objective.}
    \label{fig:intro_joint_inpaint}
    \vspace{-15pt}
\end{figure}

Current multimodal models are typically trained jointly using (an approximation to) a maximum likelihood objective over sequences consisting of images, text, and other modalities. Autoregressive (AR) models typically quantize data from continuous modalities and optimize the exact likelihood through a series of conditionals; during generation, they use a fixed token order, e.g., left-to-right, top-to-bottom (raster order) for images. They have demonstrated strong performance in both text and image generation, making them the current workhorse for multimodal models. However, generating image tokens autoregressively is slow and wasteful as nearby tokens are highly correlated, and this process results in many unnecessary forward passes through the network~\cite{lu2022unified,team2023gemini,chameleonteam2024chameleonmixedmodalearlyfusionfoundation}. Moreover, AR models are difficult to control~\cite{li2022diffusionlmimprovescontrollabletext}, cannot inpaint or infill unless explicitly trained to, and cannot easily trade-off quality versus compute at inference time.

On the other hand, continuous diffusion models—which have been shown to work well for continuous modalities such as images—have fast inference, are highly controllable, and can easily trade-off quality vs. compute. These models corrupt data by adding Gaussian noise and are trained to denoise the data, maximizing a lower bound on the likelihood. However, these models have found to be significantly slower to train in text domain compared to AR models (by roughly 64 times) ~\cite{gulrajani2024likelihood}. Text is inherently discrete, and adding continuous Gaussian noise to text token embeddings does not correspond to meaningful changes in the actual text. These trade-offs between different modeling strategies across modalities raises the question: what is the right unified generative formulation across text, image, and other modalities?

To address this, we present \model{}, a unified multimodal model based on discrete diffusion. While continuous Gaussian noise is inefficient with discrete data such as text and graphs, \model{} corrupts data with discrete noise—specifically, randomly masking tokens—and learns to map mask tokens into multimodal tokens during inference. Discrete diffusion through masking has been explored separately for generating text~\cite{Austin2021,sahoo2024simple} and images~\cite{chang2022maskgit,chang2023musetexttoimagegenerationmasked}. Such explorations have resulted in different noise schedules, transition kernels, and loss functions across the text and image domains. In this paper, we explore a discrete diffusion formulation and its applicability in jointly modeling text and image modalities with a unified set of hyperparameters.

We propose a unified architecture that jointly tokenizes text and images, and uses full self-attention to learn to map a masked token sequence to a clean token sequence by sampling from a joint vocabulary of text and image tokens. 
We evaluate \model{} across multimodal conditional and unconditional generation on multiple image-text datasets and compare to its AR counterpart. First, we find \model{} achieves a higher FID and CLIP score than AR (\cref{fig:cfg_weight_ablation}), which we attribute to the effect of classifier-free guidance. We show that \model{} exhibits strong joint image-text inpainting abilities that are not possible with prior unified generative models as seen in \cref{fig:intro_joint_inpaint}. Second, we find that \model{} consistently outperforms its AR counterpart in inference efficiency: at a given inference compute budget, our model achieves generations of higher quality and diversity (\cref{fig:inference}). Third, we show \model{} showcases stronger discriminative ability than AR on retrieval tasks due to its variable number of sampling steps (\cref{tab:retrieval}). Lastly, we scale \model{} to a 1.4B parameter model, trained on web-scale image-text datasets.


Our code, model weights, and dataset are publicly available. More qualitative visualizations are available at \url{https://unidisc.github.io}.

\section{Related Work}
\ifthenelse{\boolean{arxiv}}
{
\subsection{Unified Multi-Modal Models}
In recent years, unified models for processing multiple modalities have advanced significantly. Models like Flamingo~\cite{alayrac2022flamingo} and PaLM-E~\cite{Driess2023PaLMEAE} demonstrate strong few-shot learning capabilities across tasks. LLAVA~\cite{liu2023visualinstructiontuning} enhances LLaMa~\cite{touvron2023llama} with multimodal fine-tuning, but still uses separate encoders, limiting true unification and image generation. Recent efforts, like Perceiver IO~\cite{jaegle2021perceiver} and Unified-IO~\cite{lu2022unified}, attempt modality unification but at a smaller scale. The Chameleon project~\cite{chameleonteam2024chameleonmixedmodalearlyfusionfoundation} scales this up with a 34B parameter model trained on image-text data. However these approaches largely focus on autoregressive generation which is inefficient for high-dimensional data.

Relevant to our work, UniD3~\cite{hu2023unified} considered discrete diffusion on image and text but made several design decisions that separated each modality, using both absorbing and uniform masking, decoupling the modalities inside the model with separate operations on each. Further we couldn't compare against their model—no training code is available and were unable to reproduce their reported results using their publicly available code.

\subsection{Discrete Diffusion Models}
Discrete diffusion models have emerged as a promising alternative to continuous diffusion for discrete data types. \cite{JaschaSohl-DicksteinUnknown} introduced the first discrete diffusion model over binary variables,~\cite{HoogeboomUnknown} extended the noising process to categorical variables, demonstrating its effectiveness on image generation tasks. D3PM~\cite{Austin2021} later extended discrete diffusion to a more general set of noising processes, allowing for more flexible noise schedules. Recent work by SEDD~\cite{loudiscrete} introduced score entropy, a novel loss function for discrete diffusion models that bridges the gap between continuous and discrete spaces, and more recently, \cite{sahoo2024simple,shi2024simplified} showed text perplexity competitive with GPT-2. Most recently, \cite{nie2024scalingmaskeddiffusionmodels} looked at the scaling properties of discrete diffusion on text. While this approach shows promise for improving discrete diffusion models, these methods were primarily focused on language modeling tasks. Our work extends the application of discrete diffusion to multiple modalities and demonstrates its effectiveness in a unified architecture.
}{
\noindent \textbf{Unified Multi-Modal Models}\;\;In recent years, unified models for processing multiple modalities have advanced significantly with models such as Flamingo~\cite{alayrac2022flamingo}, LLAVA~\cite{liu2023visualinstructiontuning}, and more recently Chameleon~\cite{chameleonteam2024chameleonmixedmodalearlyfusionfoundation}. Relevant to our work, UniD3~\cite{hu2023unified} considered discrete diffusion on image and text but separated each modality, using both absorbing and uniform masking, decoupling the modalities inside the model with separate operations on each. We provide further discussion in \cref{app:related_work,sec:large_scale_quantitative}.

\noindent \textbf{Discrete Diffusion Models}\;\;First introduced by \cite{JaschaSohl-DicksteinUnknown}, discrete diffusion models, have emerged as a promising architecture for numerous tasks, particularly in the image and audio domains. More recently, discrete diffusion was extended by D3PM~\cite{Austin2021} and scaled by~\cite{sahoo2024simple,shi2024simplified}, showing text perplexity competitive with GPT-2. Our work extends the application of discrete diffusion to multiple modalities and demonstrates its effectiveness in a unified architecture. We refer to \cref{app:related_work} for an extended related work section.
}
\section{\model{}: Unified Discrete Diffusion}

\begin{figure}[t]
    \centering
    \includegraphics[width=1.0\linewidth]{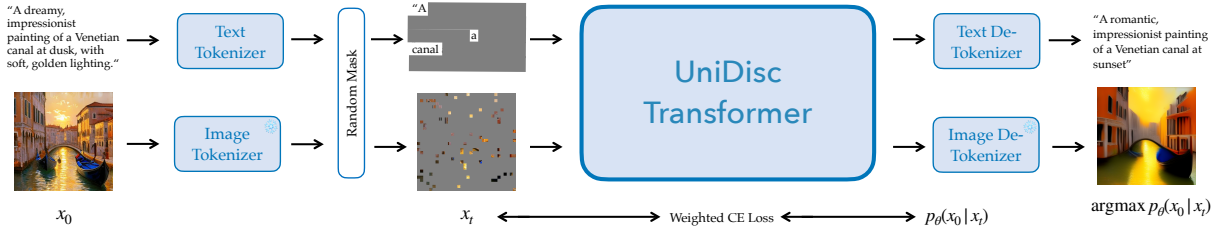}
    \caption{\model{} is a unified multimodal discrete diffusion model that can jointly process and generate text and images. Each modality is converted into a sequence of discrete tokens and we jointly denoise,  supervising with a weighted cross-entropy loss. At inference time we begin with a set of [MASK] tokens and iteratively unmask tokens.}
    \label{fig:overall_diagram}
    \vspace{-4pt}
\end{figure}

\subsection{Diffusion Models}
Diffusion models~\cite{ho2020denoisingdiffusionprobabilisticmodels,JaschaSohl-DicksteinUnknown,song2020scorebased} are a class of generative models that learn to construct a data distribution by gradually reversing a process that introduces noise into clean data samples. This approach models the transformation of a data sample \(x_0\) from a clean state through increasingly noisy states until it reaches a pure noise distribution.

The forward diffusion process is described by a series of transitions, where each latent variable \(x_t\) at time step \(t\) is sampled from a Gaussian distribution as follows:
{\setlength{\abovedisplayskip}{6pt}
\setlength{\belowdisplayskip}{6pt}
\[
q(x_t | x_0) = \mathcal{N}(x_t; \sqrt{\bar{\alpha}_t} x_0, (1 - \bar{\alpha}_t) I)
\]
}
Here, \( \bar{\alpha}_t = \prod_{s=0}^t \alpha_s \) represents the cumulative product of noise levels, making \(x_t\) increasingly distant from \(x_0\) as \(t\) increases. The variable \(x_t\) represents the noisy version of \(x_0\) at time \(t\), modeled to progressively approximate Gaussian noise as \(t\) approaches the final time step.

The reverse diffusion process then aims to reconstruct the original data by progressively denoising these samples. This involves learning the reverse transitions, with the goal to train the model \(p_\theta(x_{t-1} | x_t)\) to approximate the true reverse process and effectively recover the original data point \(x_0\) from the noisy samples.

Given $T$ timesteps of diffusion, the loss using the Evidence Lower Bound (ELBO) for the diffusion process equals\footnote{We skip the prior matching term from the loss as it is zero.}:

\vspace{-19pt}
\begin{align}
\mathcal{L}_{\text{diff}} = & \underbrace{-\mathbb{E}_{q(x_1|x_0)} \left[\log p_\theta(x_0|x_1)\right]}_{\text{reconstruction term}} \; + \label{eq:obj} \\
& \underbrace{\sum_{t=2}^T \mathbb{E}_{q(x_t|x_0)} \left[D_{KL}(q(x_{t-1}|x_t, x_0) \| p_\theta(x_{t-1}|x_t))\right]}_{\text{denoising matching term}} \notag
\end{align}
\vspace{-14pt}

\subsection{Discrete Diffusion Models}
Building on the foundations of continuous diffusion models, discrete diffusion models adapt these concepts to structures that are inherently discrete. Unlike their continuous counterparts that model transitions of $x_t$ given $x_{t-1}$ with Gaussian distributions, discrete models define transitions using categorical distributions. The forward process for discrete models is thus characterized as:

\vspace{-12pt}
\begin{equation}
q(x_t | x_0) = \text{Cat}(x_t;  x_0 \cdot \bar{Q}_t)    
\label{eq:discrete}
\end{equation}
\vspace{-16pt}

Here, \( \bar{Q}_t = \prod_{t=0}^{t=t} Q_t \)  is a $N \times N$ matrix where $N$ is the size of the vocabulary. $\bar{Q}_t$ represents the cumulative transition matrix at each discrete time step \(t\), and $Q_t$ is a transition matrix $[Q_t]_{ij} = q(x_t = j \mid x_{t-1} = i)$ dictating the probabilities of moving from one discrete state (a token in the vocabulary) \(x_{t-1}\) to another \(x_t\) discrete state (a token in the vocabulary), $x_0$ is a one-hot vector of the input data sample. D3PM~\cite{Austin2021} generalizes this framework over various transition matrices ($Q_t$), the popular ones mainly include uniform and absorbing transition matrix. In \model{}, we use the absorbing transition matrix as empirically it has been found to work the best  across text and images~\cite{Austin2021,lou2024discretediffusionmodelingestimating}.   Absorbing transition matrix requires having an absorbing state namely the [MASK] token. The matrix is represented as  $ Q_t = \alpha_tI + (1 - \alpha_t)\mathbbm{1} e_m^T$, where $\mathbbm{1}$ is a column vector of ones and $e_m$ is a one-hot vector with one on the mask state $m$. This ends up being a matrix with all zeros except $i = j \neq  m$ is $\alpha$ and $j = m, i \neq  m$ is $1- \alpha$ and $i = j = m$ is 1.

Intuitively this means that during the forward transition, the probability of an input token $x_0$ to stay the same is $\alpha$, the probability of it being masked is $1 - \alpha$, and the probability of a masked token to be unmasked is 0.

Given the forward diffusion in \cref{eq:discrete}, \cite{JaschaSohl-DicksteinUnknown} uses the same objective function as \cref{eq:obj} to optimize their model, where $q(x_{t-1} |x_t)$ ends up being a Bernoulli distribution instead of a Gaussian distribution. MDLM~\cite{sahoo2024simple} simplifies this objective function by considering continuous time-diffusion and applying loss only on the masked tokens. The final loss simply ends up being a re-weighted masked generative modeling loss:

\vspace{-15pt}
\begin{equation}
\mathcal{L}_{\text{diff}} =
\mathbb{E}_{t \sim \mathcal{U}(0,1), q(x_t \mid x)} \left[ \frac{\alpha'_t}{1 - \alpha_t} \log  p_\theta (x_0 \mid x_t) \right]
\label{eq:mdlm}
\end{equation}
\vspace{-5pt}

where $\alpha'_t = \alpha_t - \alpha_{t-1}$, and $\alpha_t$ is the probability of the token not being masked. MaskGIT~\cite{chang2022maskgit} and Muse, state-of-the-art masked image generative model use the same loss as \cref{eq:mdlm}, except there is no reweighting term and the time is discrete time instead of continuous time. The noising schedule $\alpha_t$ is also different, while language discrete diffusion models such as~\cite{Austin2021,sahoo2024simple} use a linear-time schedule, MaskGIT and Muse~\cite{chang2022maskgit,chang2023musetexttoimagegenerationmasked} use a cosine schedule. We ablate these different design choices in \cref{app:ablations}.

\subsection{Unified Training via \model{}}
We train a bidirectional decoder-only transformer~\cite{vaswani_attention_2017} using 2D RoPE embeddings~\cite{liu2024lumina} for all image tokens, 1D RoPE~\cite{su2023roformerenhancedtransformerrotary} embeddings for text tokens, and add learned modality-specific embeddings to each token. This allows our model both flexibility in resolution at inference, and the ability to use compute effectively by performing the majority of training at a lower resolution. We use the same objective function as \cref{eq:mdlm}, except for us $x_0$ is $[x_0^{img},x_0^{txt}]$

Classifier-Free guidance (CFG)~\cite{Ho2022} has been used in continuous diffusion models to trade-off between quality and diversity of generation. We apply this idea to discrete diffusion, with a probability of 0.1 we set all the tokens of a random modality to be mask tokens, this allows \model{} to learn unconditional likelihood for image and text modality. During inference we use CFG for conditional generation (image-to-text or text-to-image) to trade-off  between quality and diversity of generation as shown in \cref{fig:inference}.

\ifthenelse{\boolean{arxiv}}
{}{
Empirically, we find that image decoding requires significantly fewer sampling steps than text decoding. As can be seen in \cref{fig:inference} (c) and (d), FID saturates within 32 sampling steps for images, while it takes 400 sampling steps for the text generative perplexity to saturate. This creates an issue for unified modeling, as the number of sampling steps is bottlenecked by the maximum number of sampling steps of any input modality. To resolve this issue, we propose KV-caching the tokens of the faster modality, which in this case is the image. This however is not feasible with current training scheme, as the text tokens never encounter the image tokens from the previous denoising steps.

To fix this we propose to have different time schedules for each modality: specifically, a slower time-schedule for text and a faster one for images. To implement these schedules, we consider $N_{min}$ and $N_{max}$, which represent the min and max number of timesteps one might consider during inference. We also consider $K$, which represents the text-to-image inference ratio, i.e how many times is image inference faster than text inference. Empirically, we find this number to be roughly 10, as seen in \cref{fig:img_txt_ratio}. During training, we first randomly sample text timesteps from a uniform distribution, $t_{txt} \sim \mathcal{U}(0,1)$ and set image timesteps by randomly offsetting the text timesteps with $\delta t_i$, specifically, $t_{img} \sim \mathcal{U}(t_{txt}, t_{txt} - {\delta t}_i)$. This ensures that the image timestep only moves behind the text timestep by a maximum of ${\delta t}_i$.
We randomly set ${\delta t}_i$ to $ {\delta t}_i \sim  \mathcal{U}(\frac{K}{N_{max}}, \frac{K}{N_{min}})$. This modality-specific timestep schedule ensures \model{} can handle image token KV caching during inference, resulting in faster inference speeds as seen in \cref{fig:cache_time}.
}

To improve training stability, we use Query-Key Normalization~\cite{wortsman2023small} and use RMSNorm~\cite{zhang2019root} for all other norms. We use Sandwich Normalization—normalization before and after each FFN, as we found this helps control activations in deeper layers as previously reported in~\cite{DBLP:conf/nips/DingYHZZYLZSYT21,zhuo2024lumina}. 

To further improve the convergence speed of discrete diffusion we analyze the noising schedule and find that linear schedule in~\cite{sahoo2024simple,Austin2021} results in excessively high weighting for early timesteps, impairing the convergence speed. Following Min-SNR trick in continuous diffusion~\cite{hang2023efficient}, we limit the minimum weighting to 5. An architecture diagram is provided in \cref{fig:overall_diagram} and pseudo-code for training procedure is provided in \cref{app:training_algo}.

\subsection{Unified Sampling via \model{}}
Sampling in masked discrete diffusion, involves mapping a set of masked tokens $m$ to a set of visible tokens $x_0$ using $T$ timesteps of denoising. A variety of sampling strategies have been previously proposed~\cite{JaschaSohl-DicksteinUnknown, Austin2021,zheng2024masked,chang2022maskgit,sahoo2024simple,lou2024discretediffusionmodelingestimating} for masked discrete diffusion. MaskGIT~\cite{chang2022maskgit} proposes a confidence-based sampling, where they decode the most confident tokens at each step of denoising. D3PM~\cite{Austin2021} and MLDM~\cite{sahoo2024simple} uses a sampling mechanism similar to~\cite{ho2020denoisingdiffusionprobabilisticmodels} except applied to a bernoulli distribution, which we refer to as DDPM sampling. This results in a random set of tokens being decoded, instead of the most confident ones as in MaskGIT. We ablate these sampling strategies in \cref{fig:inference} and find the confidence-based sampling proposed in MaskGIT to work the best for unified modeling.

\ifthenelse{\boolean{arxiv}}
{}{
We also build on top of MaskGIT sampling and add modality-specific caching and nucleus sampling on the logits. In modality-specific caching, we cache the image token keys/values every $k$ steps. We provide the pseudo-code of our sampling in Algorithm \ref{alg:sampling}. In \cref{fig:cache_time}, we find that KV caching the image tokens results in much lower latency at higher sequence length or batch size. Our algorithm along with our MaskGIT implementation is available in \ref{app:sampling_algo}.
}

\begin{figure*}[t]
    \centering
    \begin{minipage}[t]{0.48\linewidth}
        \centering
        \includegraphics[width=\linewidth]{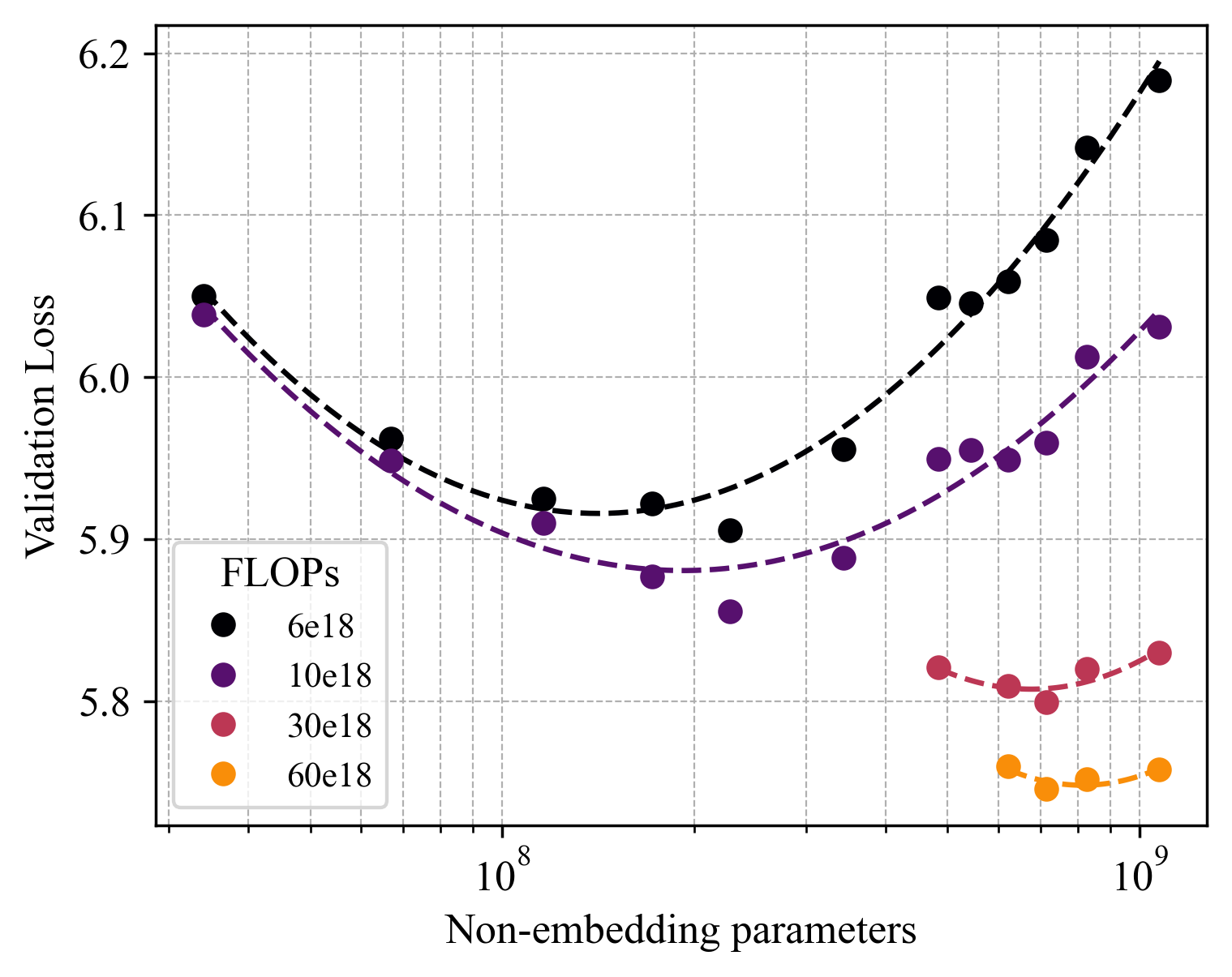}
    \end{minipage}
    \hfill
    \begin{minipage}[t]{0.48\linewidth}
        \centering
        \includegraphics[width=\linewidth]{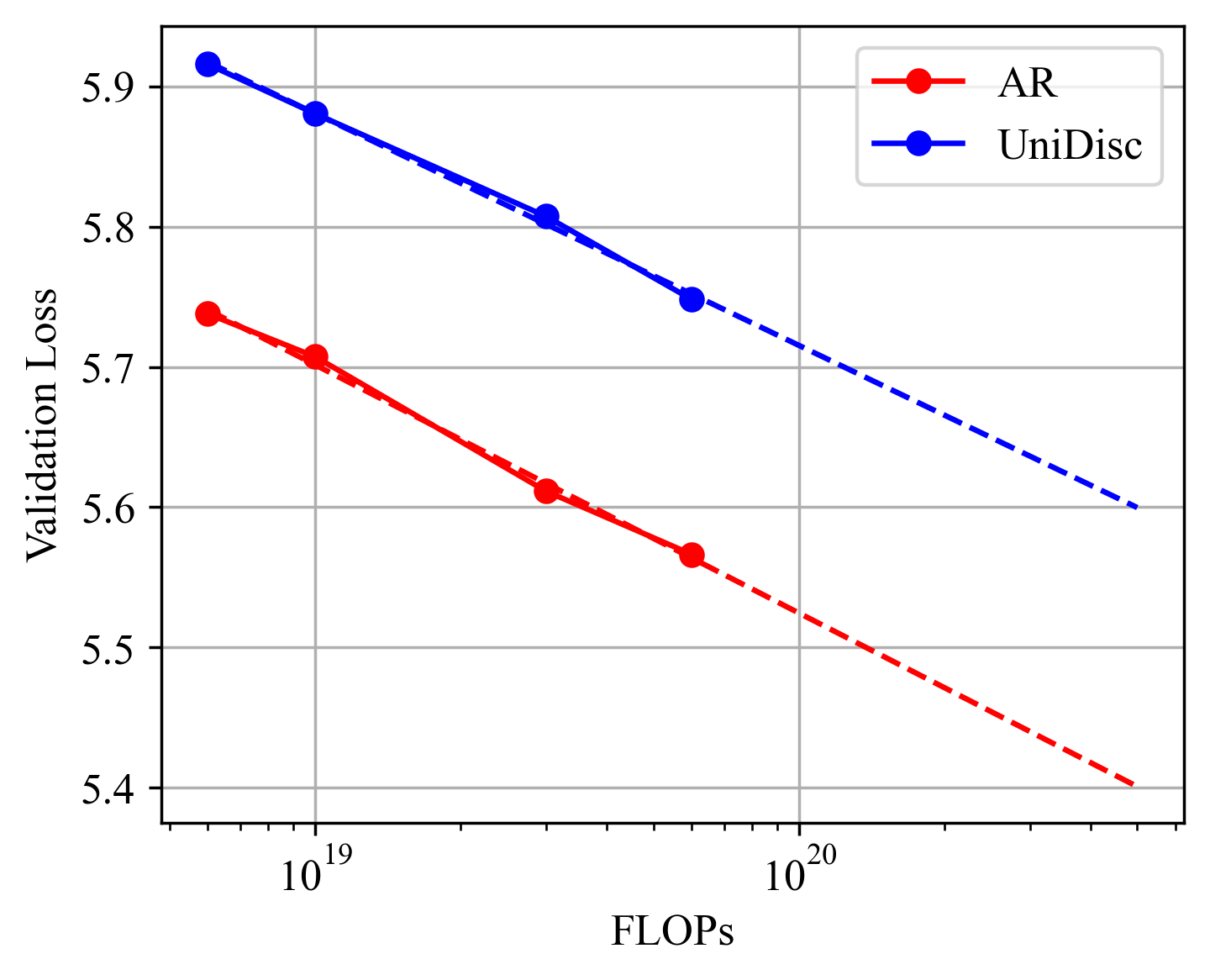}
    \end{minipage}
    \vspace{-8pt}
    \caption{\textbf{Scaling Analysis for AR and \model{} models}: (Left) IsoFLOP curves for \model{}, plotting varying model size for a fixed FLOP budget. (Right) Estimating optimal parameter size for each budget - minima of fitted parabola, we plot scaling laws for both AR and \model{}. We find 13.2x more compute is required for \model{} to achieve the same overall loss as AR.}
    \label{fig:scaling_laws}
    \vspace{-6pt}
\end{figure*}

\section{Experiments}
\label{experiments}

We compare \model{} against an autoregressive (AR) baseline across various tasks, metrics and datasets. We use the same architecture and hyper-parameters, and data, only differing in the attention mask and respective loss functions. For our autoregressive baseline we use a standard language model architecture from Chameleon~\cite{chameleonteam2024chameleonmixedmodalearlyfusionfoundation}—that is a decoder-only transformer with causal attention and rotary positional embeddings. To enable classifier-free guidance, we dropout modalities with $10\%$ probability during training. For \model{}, we dropout both modalities and for the AR baseline we dropout only the first modality in the input sequence as in~\cite{liu2024lumina}. 

Our experiments aim to answer the following questions:
\vspace{-7pt}
\begin{enumerate}[itemsep=-2pt]
\item How does \model{} compare against AR models in unconditional and conditional multi-modal generation of image/text pairs?
\item How effective is classifier-free guidance in conditional generation for AR models and for \model{}?
\item How does \model{} compare against AR models in terms of training efficiency with varying the ratio of image-text tokens?
\item How do various sampling strategies for \model{} affect its generation results and inference speed?
\item How does \model{} compare against AR models across image-language reasoning tasks?
\item How do various design choices of \model{} contribute to its performance?
\end{enumerate}
\vspace{-4pt}
Lastly, we show that we can successfully scale \model{}, to a 1.4B parameter model, trained on 500B tokens. We qualitatively evaluate this model, to demonstrate its capabilities.

\textbf{Datasets:} In \cref{sec:uncond_cond_exp}, \ref{ssec:efficiency}, and~\ref{img_text_retr}, we conduct experiments with different train and validation sets. Our training set includes DataComp1B~\cite{gadre2024datacomp}, CC12M~\cite{changpinyo2021conceptual12mpushingwebscale}, CLEVR-math~\cite{clevrmath}, and CLEVR-Ref~\cite{liu2019clevrrefdiagnosingvisualreasoning}. Our evaluation datasets include a held-out validation set of DataComp1B and CC12M, along with Flickr~\cite{plummer2016flickr30kentitiescollectingregiontophrase}, MS-COCO30k~\cite{chen2015microsoftcococaptionsdata}, and Winoground~\cite{thrush2022winogroundprobingvisionlanguage}. 

\subsection{Evaluation of Multimodal Generation}
\label{sec:uncond_cond_exp}
We evaluate \model{} and AR models in unconditional and conditional generation tasks.

\begin{figure*}[t]
    \centering
    \ifthenelse{\boolean{arxiv}}
    {
    \begin{minipage}{0.51\columnwidth}
        \hspace*{-0.1\columnwidth}
        \includegraphics[width=\linewidth]{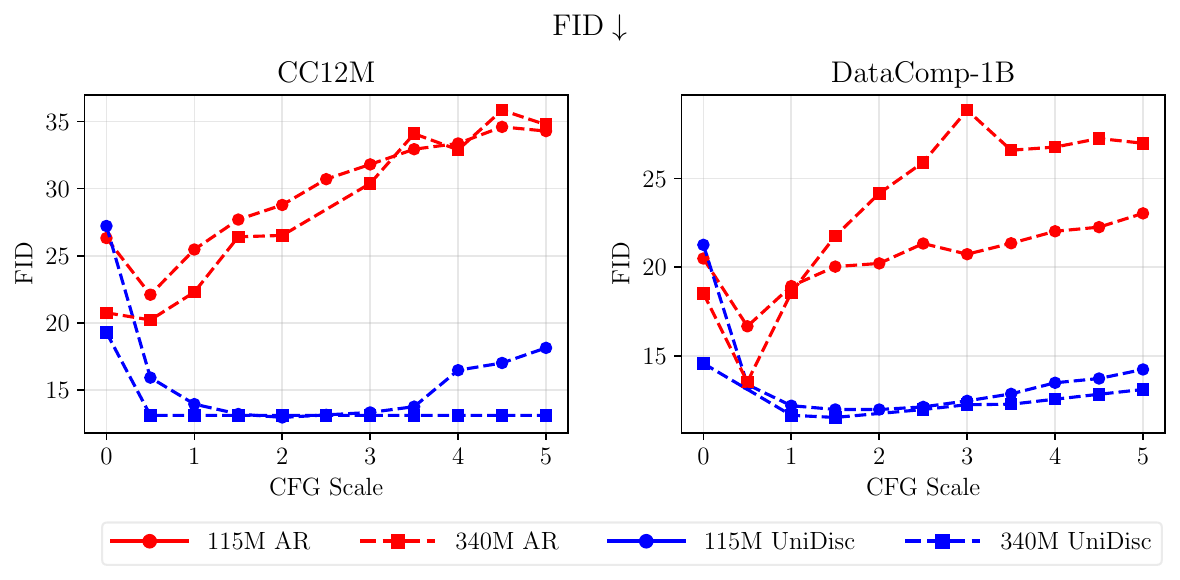}
    \end{minipage}%
    \begin{minipage}{0.51\columnwidth}
        \includegraphics[width=\linewidth]{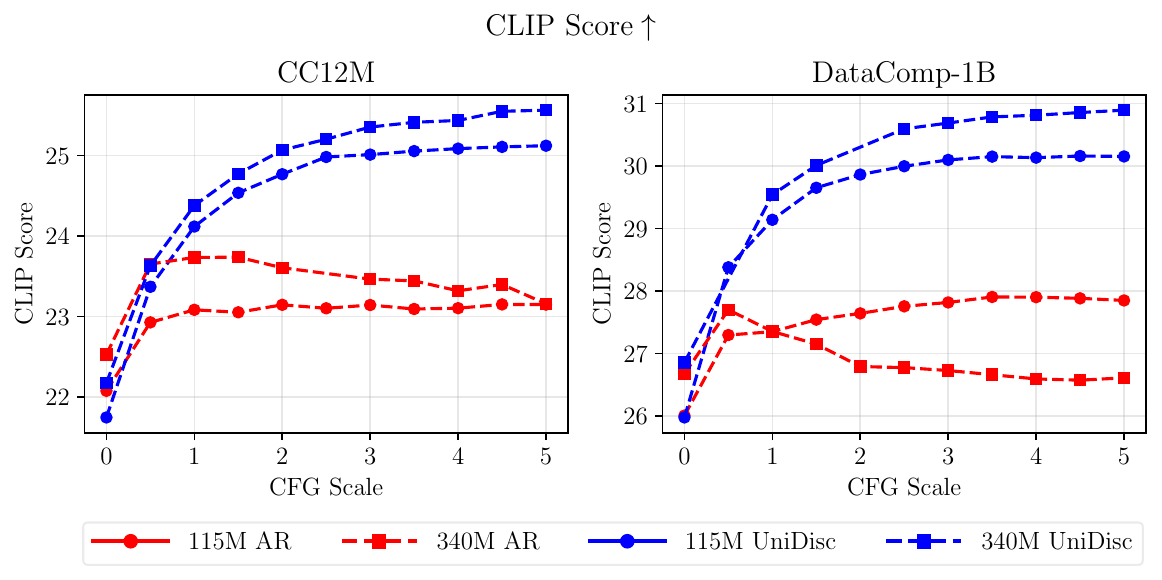}
        \hspace*{-0.1\columnwidth}
    \end{minipage}
    }
    {
    \begin{minipage}{1.05\columnwidth}
        \hspace*{-0.05\columnwidth}
        \includegraphics[width=\linewidth]{images/cfg_fid_top_half.pdf}
    \end{minipage}%
    \begin{minipage}{1.05\columnwidth}
        \includegraphics[width=\linewidth]{images/cfg_clip_score_top_half.pdf}
        \hspace*{-0.05\columnwidth}
    \end{minipage}
    }
    \vspace{-4pt}
    \caption{\textbf{Conditional generation results for both FID and CLIP metrics, across a range of CFG values.} We find that AR is more sensitive to the CFG weighting, with a narrower optimal range.}
    \label{fig:cfg_weight_ablation}
\end{figure*}


\noindent \textbf{Evaluation metrics:}
We consider the following three evaluation metrics, most commonly used in previous works: \textbf{i) Joint perplexity}  indicates a model's ability to fit to different validation sets. Note that this metric is jointly calculated across image-text tokens. The perplexity values from the autoregressive  Chameleon baseline are exact likelihoods, the values from \model{} are upper bounds. While perplexity is a good metric for assessing the  fitting ability of a model, it cannot be used to evaluate its generation ability. \textbf{ii) Fréchet inception distance (FID)}~\cite{heusel2017gans} is a popular metric in image-generation to quantify the quality and diversity of image generation.
\textbf{iii) CLIP score} is used for calculating image-text coherence. While we could not find an equivalent to the FID metric for text, we use CLIP score to evaluate generated image-text coherence, conditioning our model on an input image. 

\noindent \textbf{Experimental details:}
\label{exp_details}

We show conditional image-text generation results in  \cref{fig:cfg_weight_ablation}.  We condition on an image to generate the corresponding language description, and vice versa, condition on the language description to generate the corresponding image.  For unconditional results please refer to \cref{tab:uncond} in the Appendix.  
We use a dataset comprising $30$M image-text pairs from DataComp1B~\cite{gadre2024datacomp} and CC12M~\cite{changpinyo2021conceptual12mpushingwebscale}, please refer to \cref{app:cond_uncond_details} for further details.

\textit{\model{} significantly outperforms AR in conditional generation}  \textit{while performing equally well in unconditional generation}). We attribute this performance gap to classifier-free guidance (CFG). As can be seen in \cref{fig:cfg_weight_ablation}, the results without CFG ($\operatorname{Scale}\!=\!0$) are similar between AR and \model{}, but increasing CFG disproportionately benefits \model{}.

The iterative generation process of diffusion makes it easy to blend conditional and unconditional predictions to guide the output. Autoregressive models, on the other hand, generate data sequentially in a fixed order, without any iterative refinement, which makes it difficult to mix in unconditional predictions to guide generation. We study this in detail in Section \ref{sec:cfg_study} in Appendix.

\vspace{-3pt}
\paragraph{Joint Image-Text Inpainting} In \cref{fig:intro_joint_inpaint}, we show that \model{} can inpaint in a joint text and image space—without any fine-tuning. Currently, none of the popular generative models have this capability, because most multimodal generative models are either autoregressive~\cite{ChameleonTeam2024} or use mixed modeling~\cite{Zhou2024}, which prevents them from easily inpainting jointly over image and text. In \cref{ssec:inpainting_ar_nar} we explicitly fine-tune an AR model for joint inpainting, \model{} zero-shot still shows far better performance. For more qualitative results, please refer to \cref{fig:text_cond_image_inpaint}, and \cref{fig:appendix_joint_image_inpaint} in the Appendix.

\ifthenelse{\boolean{arxiv}}{}{
\begin{figure}[tb]
    \centering
    \includegraphics[width=\linewidth]{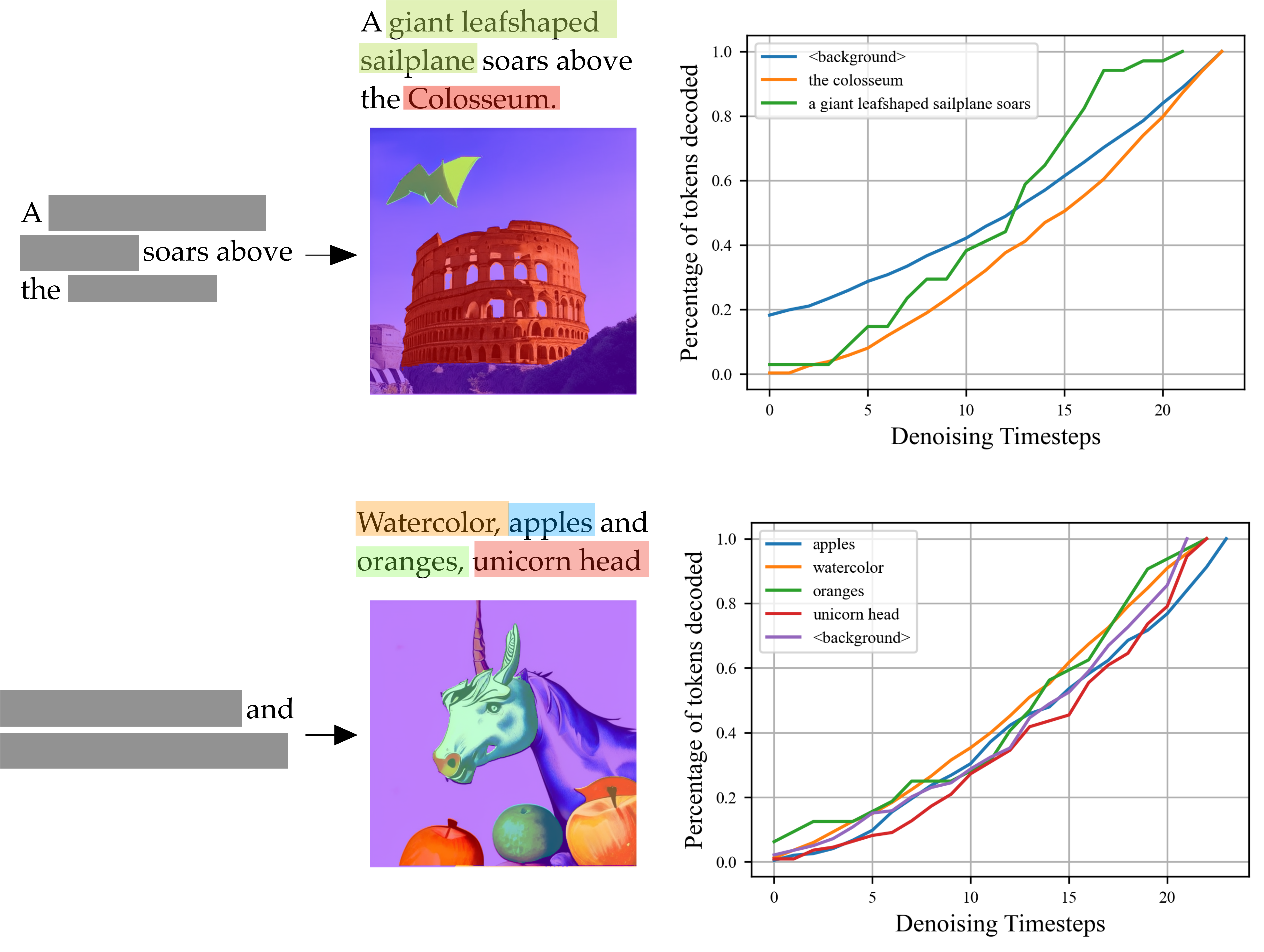}
    \vspace{-16pt}
    \caption{\textbf{Uniform Concept Generation}: We perform joint generation given only masked text input (left). We use a language-conditioned segmentation model and find that \model{} generates uniformly in \textit{concept space} (right).}
    \label{fig:seg_viz_small}
    \vspace{-2pt}
\end{figure}
}

\ifthenelse{\boolean{arxiv}}{}{
\begin{figure}[tb]
    \centering
    \ifthenelse{\boolean{arxiv}}{
    \includegraphics[width=\linewidth]{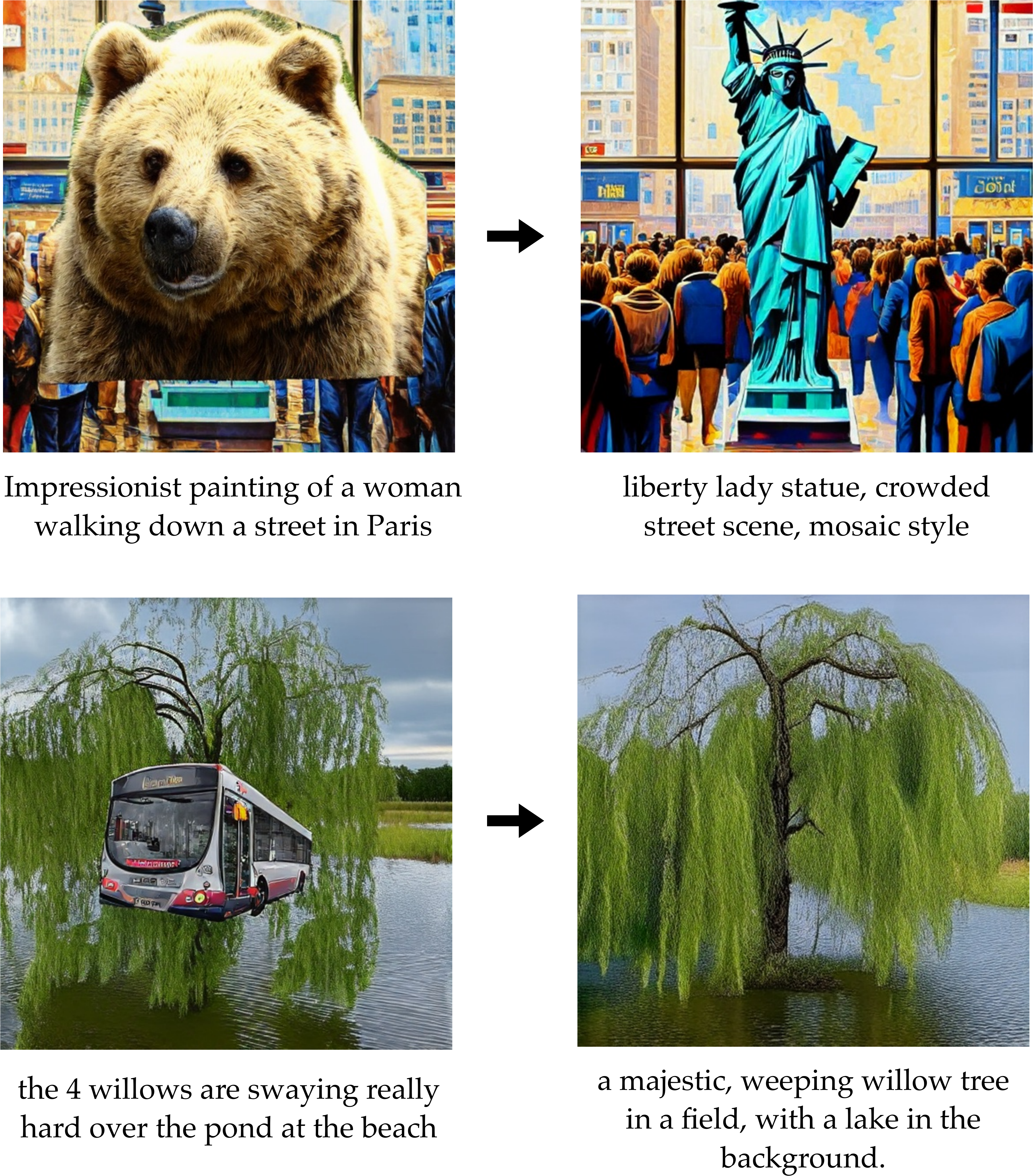}
    }{
    \includegraphics[width=\linewidth]{images/automatic_editing_small_v2.pdf}
    }
    \vspace{-16pt}
    \caption{\textbf{Zero-shot Image Editing}: \model{} can take corrupted and mismatched image/text pairs (left) and produce an aligned, high-quality pair (right), using the model's own likelihood as a scoring function.}
    \label{fig:automatic_editing_small}
    \vspace{-6pt}
\end{figure}
}

\vspace{-5pt}
\subsection{Training Efficiency and Inference Speed}
\label{ssec:efficiency}

\ifthenelse{\boolean{arxiv}}{
\begin{figure}[t!]
    \centering
    \begin{adjustbox}{width=0.9\linewidth, clip, trim=6.5mm 0 0 0}
    \includegraphics{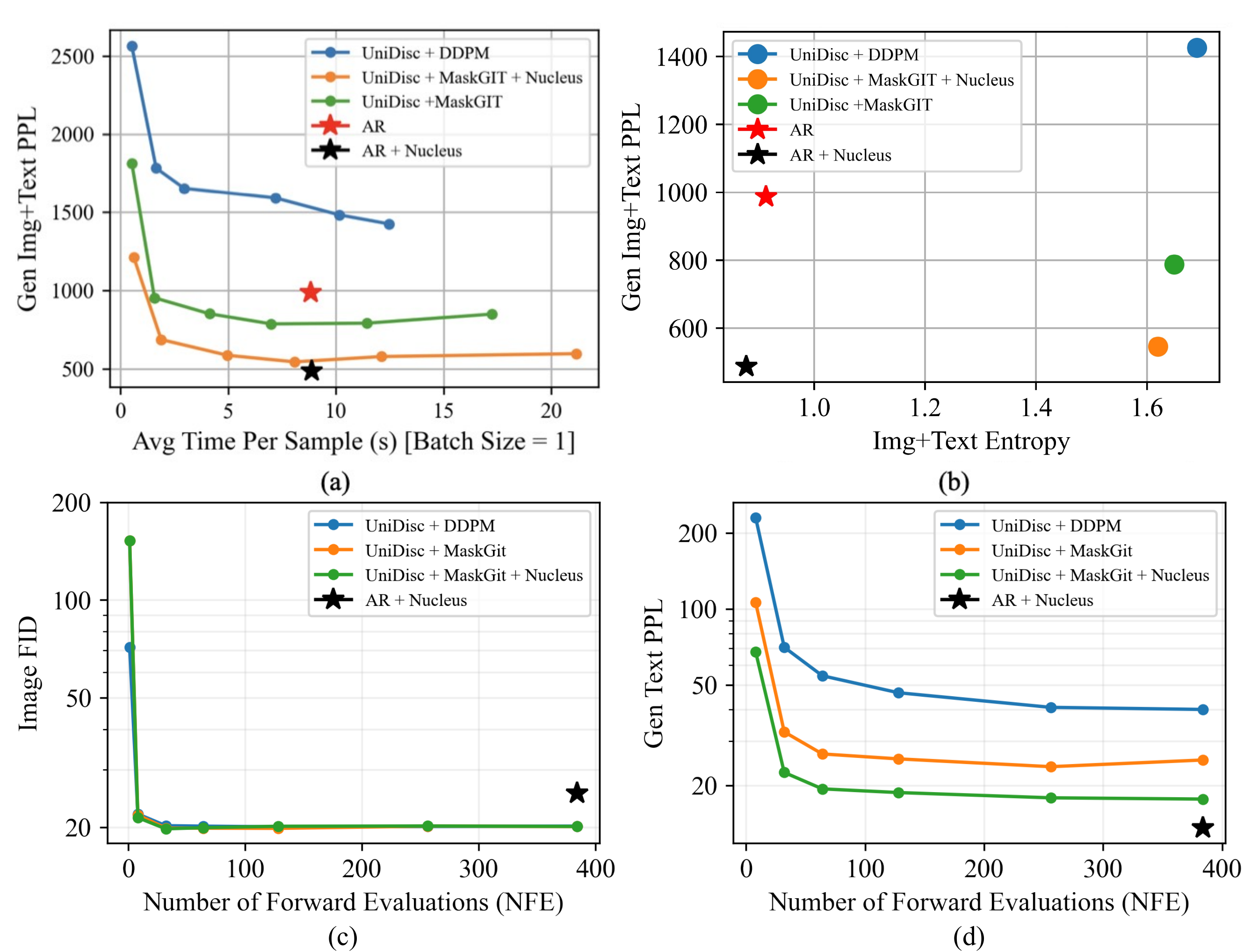}
    \end{adjustbox}
    \vspace{-4pt}
    \caption{\textbf{Inference Comparisons for \model{} and AR baseline}: (a) Chameleon Perplexity (Text+Image) vs. Time - we perform similar to best AR method, (b) Chameleon Perplexity vs. Entropy - \model{} has high diversity and low perplexity, while AR has significantly lower diversity, (c) Image FID vs. NFE, showing image generation saturates quickly with NFE ($\approx32$), (d) GPT2 Generative Text Perplexity vs. NFE showing text generation benefits from more sampling steps (diminishing).}
    \label{fig:inference}
    \vspace{-15pt}
\end{figure}
}{
\begin{figure}[tb]
    \centering
    \begin{adjustbox}{width=1.03\linewidth, clip, trim=6.5mm 0 0 0}
    \includegraphics{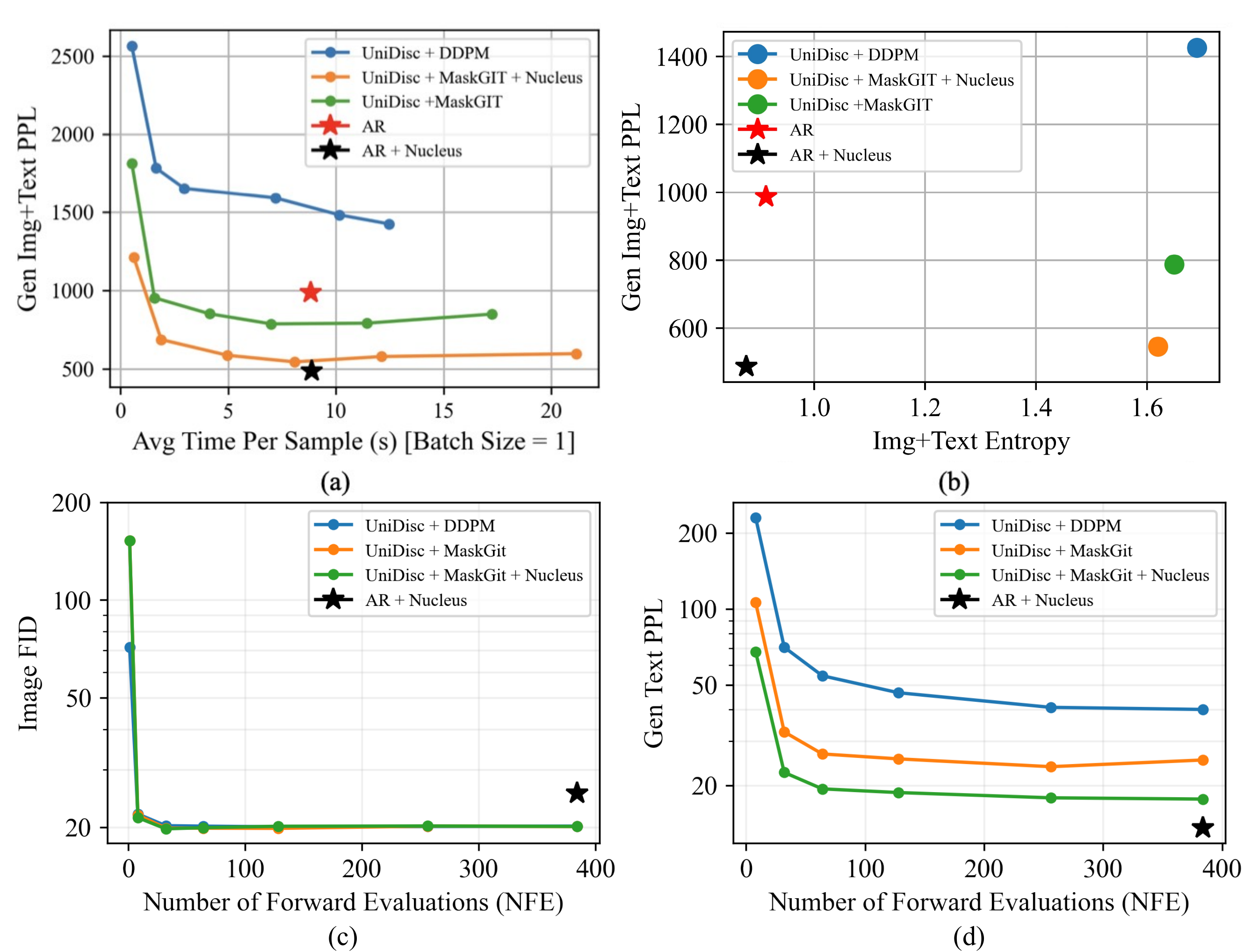}
    \end{adjustbox}
    \vspace{-18pt}
    \caption{\textbf{Inference Comparisons for \model{} and AR baseline}: (a) Chameleon Perplexity (Text+Image) vs. Time - we perform similar to best AR method, (b) Chameleon Perplexity vs. Entropy - \model{} has high diversity and low perplexity, while AR has significantly lower diversity, (c) Image FID vs. NFE, showing image generation saturates quickly with NFE ($\approx32$), (d) GPT2 Generative Text Perplexity vs. NFE showing text generation benefits from more sampling steps (diminishing).}
    \label{fig:inference}
    \vspace{-7pt}
\end{figure}
}

With the ever growing scale of recent generative models, an important aspect of their performance is their compute efficiency. Prior works~\cite{ScalingLawsNeuralLanguageModels,hoffmann2022trainingcomputeoptimallargelanguage} have extensively measured the training scaling laws of AR models, finding a power-law relationship between compute cost and distribution fitting, measured by negative log likelihood (NLL). In contrast, there has been little work that has measured the training efficiency of discrete diffusion models: the closest work is that of~\cite{gulrajani2024likelihood}, which finds that the training efficiency of continuous diffusion models is approximately 64x worse than AR models on text. Recently, concurrent work~\cite{zheng2024masked} studied discrete diffusion models, again only on text, and found a scaling factor of 16x compared to AR models.

Although discrete diffusion is thought to be comparatively more efficient on other modalities such as images, we are not aware of prior work that has empirically measured this. We perform an ISOFlop analysis~\cite{hoffmann2022trainingcomputeoptimallargelanguage} of \model{} and our AR baseline, changing only the attention mask and loss function. As in prior work, we select a set of compute budgets $C_i$ and, within each budget, vary the non-embedding parameters (incl. LM head) $N$, and total tokens during training $D$, keeping the total compute, measured in FLOPs, fixed using an approximation of $C\approx 6ND$.

We compare the training efficiency of \model{} and our AR baseline in  Figure \ref{fig:scaling_laws} (right) and find that the rough training-inefficiency factor for discrete diffusion to that of AR models for unified training is about $13.2$—i.e. one needs to train \model{} 13.2x longer to achieve the same loss. Additional experimental details are available in \cref{app:scaling}.


While training efficiency is important, inference efficiency is equally—if not more—important as we deploy these models at wide scale. Thus, we compare the inference efficiency of \model{} and our AR baseline in Figure~\ref{fig:inference} (a), (c) and (d). In (a), we measure the joint generative perplexity using Chameleon, In (c) we measure the Image FID and in (d) we measure the Text Perplexity. While it might appear from (a) and (d) that AR does better than \model{}, in Figure \ref{fig:inference} (b), we find that \model{} has far higher entropy at a given perplexity.

We note that solely looking at the generative perplexity is not sufficient, as it has been previously found~\cite{zheng2024masked} that very low perplexity can be achieved by repeating the same tokens, which we find often happens with AR w/nucleus sampling and low temperature. We demonstrate such degenerate cases in \cref{app:low_pplx_examples}. Therefore, Generative Perplexity + Entropy should be considered jointly for evaluating the quality of generation results.

\subsection{Multimodal Discriminative Performance}
\label{img_text_retr}

Generative models can act as strong discriminative models as shown in several recent works~\cite{li2023diffusionmodelsecretlyzeroshot,jaini2024intriguingpropertiesgenerativeclassifiers,prabhudesai2023diffusionttatesttimeadaptationdiscriminative}. Moreover,~\cite{rambhatla2023selfevalleveragingdiscriminativenature} show the discriminative ability of a generative model can be a good metric for its generation performance. In this section, we compare the discriminative capabilities of AR models and \model{} on cross-model retrieval tasks (image/text/joint).

We evaluate on Winoground~\cite{thrush2022winogroundprobingvisionlanguage} and a held-out DataComp1B validation set~\cite{gadre2024datacomp}, using $18$M text/image pairs from DataComp1B as our training set. To enable text retrieval during inference for the AR model, we train with flipping the order of modalities, putting the image first $20\%$ of the time, following~\cite{Zhou2024}. We find that this improves the retrieval for the AR model. All other hyperparameters follow those in \cref{sec:uncond_cond_exp}. Details of evaluations on CLEVR-VQA and CLEVR-Ref are available in \cref{ssec:discrim_evaluations}

\ifthenelse{\boolean{arxiv}}{
\begin{figure*}[h!]
    \centering
    \begin{minipage}{0.48\columnwidth}
        \centering
        \includegraphics[width=\linewidth]{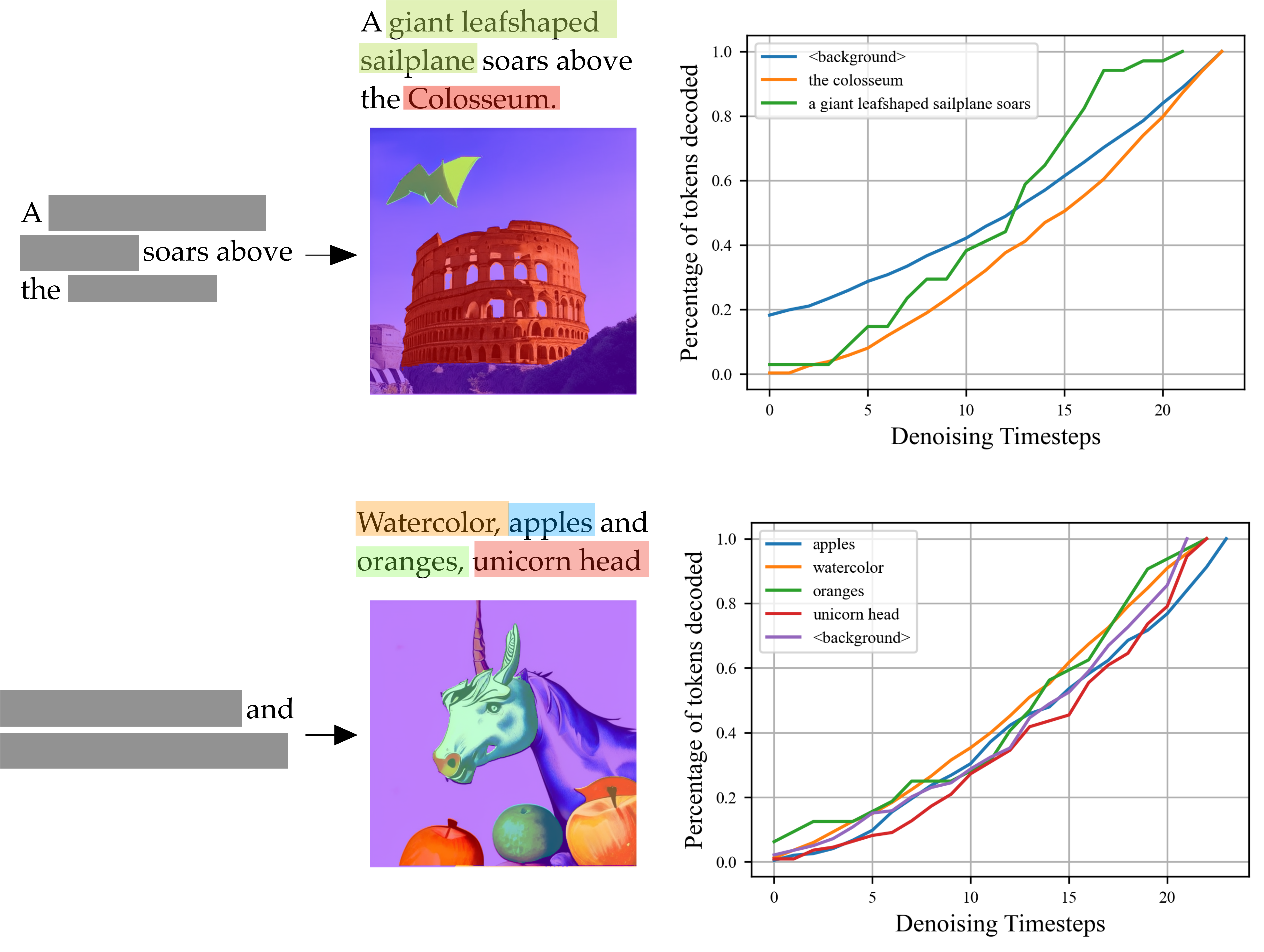}
        \vspace{-16pt}
        \caption{\textbf{Uniform Concept Generation}: We perform joint generation given only masked text input (left). We use a language conditioned segmentation model and find that \model{} generates uniformly in \textit{concept space} (right).}
        \label{fig:seg_viz_small}
        \vspace{-2pt}
    \end{minipage}
    \hspace*{0.02\columnwidth}
    \begin{minipage}{0.48\columnwidth}
        \centering
        \renewcommand{\arraystretch}{1.5}
        \resizebox{0.95\linewidth}{!}{
        \begin{tabu}{lccccccccccccccccccccc}
            & \rotatebox{0}{Clevr-VQA} & \rotatebox{0}{Clevr-Ret} & \rotatebox{0}{Datacomp}  & \rotatebox{0}{Winoground}   \\
            \midrule
            \textbf{Text Retrieval}  \\
            AR & 0.60 & 0.81 & 0.85 &  0.24  \\
            \hdashline
            \model{} & \textbf{0.63} & \textbf{0.94} & 0.85  &\textbf{0.31} \\
            \midrule
            \textbf{Image Retrieval}  \\
            AR & N/A & 0.06 & \textbf{0.96}  & 0.25 \\
            \hdashline
            \model{} & N/A & \textbf{0.25} & 0.95  & \textbf{0.27}  \\
            \midrule    
            \textbf{Joint Retrieval}  \\
            AR & N/A & 0.06 & 0.17  & 0.06  \\
            \hdashline
            \model{} & N/A & \textbf{0.5} & \textbf{0.64} & \textbf{0.20} \\
            \bottomrule    
            \end{tabu}
        }
        \caption{\textbf{Image-Text Reasoning measured by QA and retrieval accuracy across datasets.}}
        \label{tab:retrieval}
    \end{minipage}
\end{figure*}
}{
\begin{table}[t]
    \centering
    \renewcommand{\arraystretch}{1.5}
    \resizebox{0.95\linewidth}{!}{
    \begin{tabu}{lccccccccccccccccccccc}
        & \rotatebox{0}{Clevr-VQA} & \rotatebox{0}{Clevr-Ret} & \rotatebox{0}{Datacomp}  & \rotatebox{0}{Winoground}   \\
        \midrule
        \textbf{Text Retrieval}  \\
        AR & 0.60 & 0.81 & 0.85 &  0.24  \\
        \hdashline
        \model{} & \textbf{0.63} & \textbf{0.94} & 0.85  &\textbf{0.31} \\
        \midrule
        \textbf{Image Retrieval}  \\
        AR & N/A & 0.06 & \textbf{0.96}  & 0.25 \\
        \hdashline
        \model{} & N/A & \textbf{0.25} & 0.95  & \textbf{0.27}  \\
        \midrule    
        \textbf{Joint Retrieval}  \\
        AR & N/A & 0.06 & 0.17  & 0.06  \\
        \hdashline
        \model{} & N/A & \textbf{0.5} & \textbf{0.64} & \textbf{0.20} \\
        \bottomrule    
        \end{tabu}
    }
    \caption{\textbf{Image-Text Reasoning measured by QA and retrieval accuracy across datasets.}}
    \label{tab:retrieval}
    \vspace{-4pt}
\end{table}
}

\ifthenelse{\boolean{arxiv}}{

    
}{
\begin{figure}[t]
    \centering
    \includegraphics[width=\columnwidth]{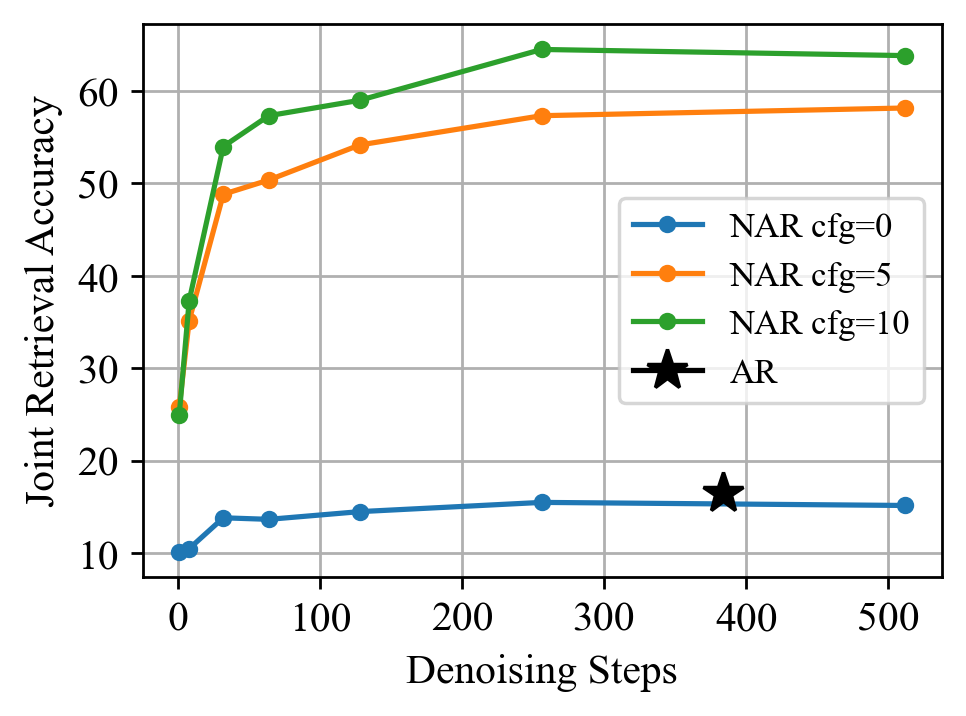}
    \vspace{-10pt}
    \caption{\textbf{Joint Retrieval Accuracy on DataComp1B.} We outperform AR given the task of retrieving one correct image-text pair out of 16 possible pairs, implying better learnt representations.}
    \label{fig:retrieve}
    \vspace{-2pt}
\end{figure}
}

In \cref{tab:retrieval}, we report the image retrieval, text retrieval and joint retrieval accuracy for AR and \model{}. For image retrieval, the model is given a text caption paired with 16 images,  out of which only one image is correctly paired and the rest are random. The goal is to accurately classify the correct image. To evaluate the model's retrieval accuracy we check if the correct image has the highest  $ p(x^{img} | x^{txt})$ among all other images. We do the same for text retrieval, where we check $ p(x^{txt} | x^{img})$. For joint retrieval, only a single pair has the correct mapping, and every other pair has a random image and text. We check if the correct pair has the highest joint probability $p(x_{img},x_{txt})$

We find that \model{} significantly outperforms the AR model on all retrieval tasks. To further investigate this, we measure the joint retrieval accuracy across denoising steps \& CFG values in \cref{fig:retrieve} in Appendix. We find CFG and the number of denoising steps to play a large role in \model{}'s retrieval accuracy. While the number of denoising steps in an AR model is fixed to the sequence length, the denoising steps for \model{} can be much higher.

\newpage
\subsection{Scaling \model{}}
\label{sec:large_scale}
\begin{wrapfigure}{r}{0.48\columnwidth}
    \centering
    \ifthenelse{\boolean{arxiv}}{
    \includegraphics[width=\linewidth]{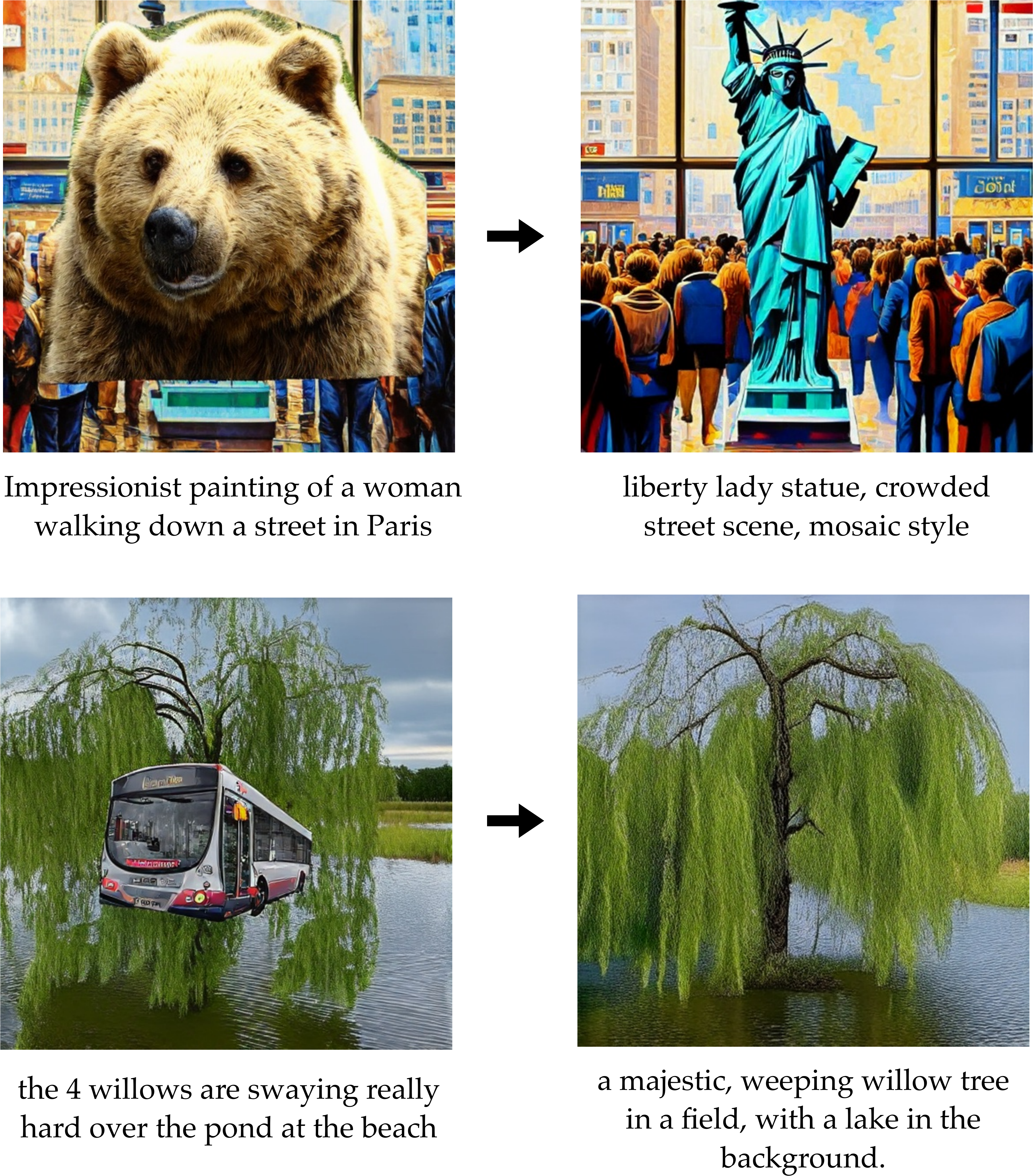}
    }{
    \includegraphics[width=\linewidth]{images/automatic_editing_small_v2.pdf}
    }
    \vspace{-16pt}
    \caption{\textbf{Zero-shot Image Editing}: \model{} can take corrupted and mismatched image/text pairs (left) and produce an aligned, high-quality pair (right), using the model's own likelihood as a scoring function.}
    \label{fig:automatic_editing_small}
    \vspace{-6pt}
\end{wrapfigure}
We show that \model{} scales well across parameters and dataset size. We train a 1.4B parameter model with web-scale data. Our model is trained in two stages, with a low-resolution pre-training stage and a second high-resolution fine-tuning stage. Our first-stage consists of $250$M image/caption pairs at 256$\times$256 resolution. We curate our dataset from several sources, with $200$M open-web images from~\cite{gadre2024datacomp}, which were re-captioned by a VLM to create higher-quality descriptions by~\cite{li2024if}. We also add a set of smaller datasets consisting of PixelProse~\cite{singla2024pixelsproselargedataset}, JourneyDB~\cite{sun2023journeydbbenchmarkgenerativeimage}, and Cambrian-10M~\cite{tong2024cambrian}. In addition, we construct a high-quality, custom dataset of $18$M synthetic images, following findings by~\cite{zhuo2024lumina,sehwag2024stretching} on the importance of image/caption alignment for image generation. We construct our dataset by prompting an LLM to augment a set of 250K human prompts and use~\cite{esser2024scaling} for generations. In both stages, we account for dataset imbalance and sample more from higher-quality sources. Finally, we fine-tune our model in a second stage, interpolating the RoPE 2D embeddings to train at 512$\times$512 on $30$M image/caption pairs.

Further due to lack of space, we ablate several architecture and objective design choices on a smaller model in \cref{app:ablations} and we show the training curve of our 1.4B model is available in \cref{fig:finetune_large_scale_curve}. We also compare \model{} to recent multimodal models on standard image generation benchmarks in \cref{sec:large_scale_quantitative}.  Qualitative results from the model are available in \cref{app:large_scale_extra_qual}, demonstrating zero-shot text-conditioned image inpainting (\cref{fig:text_cond_image_inpaint}), standard text-to-image (\cref{fig:t2i}), and image-to-text (\cref{fig:i2t}) generation. Moreover, we demonstrate a form of image editing in \cref{fig:automatic_editing_small} and ~\cref{ssec:automatic_image_editing}, showing that \model{} can, without any specialized fine-tuning, \textit{automatically} improve a text \& image pair by noising and denoising, using the model's likelihood as a judge. Additionally, we analyze the joint generation of \model{} in \cref{fig:seg_viz_small} and \cref{ssec:joint_generation_analysis} , finding that the model generates images roughly \textit{uniformly in concepts instead of in area}.

\section{Conclusion}
In this paper, we introduced \model{}, the first large-scale unified multimodal discrete diffusion model capable of generating, inpainting and editing both images and text. By leveraging discrete diffusion processes, we showed that \model{} surpasses autoregressive models in both inference efficiency and quality. Our model unifies various design choices in discrete diffusion space, across modalities, through extensive ablations and analysis. We hope that our work inspires future research in this direction.

\newpage
\section{Acknowledgment}
This work is funded by LambdaLabs, ONR award N00014-23-1-2415, AFOSR FA9550-23-1-0257, AFOSR FA9550-23-1-0747, ONR MURI N00014-24-1-2748, ONR MURI N00014-22-1-2773 and DARPA No. HR00112490375 from the U.S. DARPA Friction for Accountability in Conversational Transactions (FACT) program.  We thank CMU FLAME Cluster, LambdaLabs and Google TPU Research cloud for providing us with compute for this work. We also thank Alex Li, Unnat Jain and Lili Chen for helpful discussion in regards to the project.

\bibliography{references,refs,zotero}
\bibliographystyle{abbrvnat}

\newpage
\appendix
\onecolumn
\makeatletter
\renewcommand{\@makecaption}[2]{%
  \vskip 10pt
  \centering
  {#1. #2}
}
\makeatother

\section{\model{}}


\subsection{\model{} Training}
\label{app:training_algo}
We describe the detailed algorithm for unified discrete diffusion training on image and text below in Algorithm \ref{alg:training}.

\begin{algorithm}[H]
\caption{\model{} Training}
\begin{algorithmic}[1]
\State \textbf{Require:} Training data $x$
\State \textbf{Require:} Noising Schedule $\alpha_t$ i.e., Linear or Cosine
\State \textbf{Require:} Unconditional probability $p_{uncond}$
\State \textbf{Initialize:} Model parameters $\theta$
\Repeat
    \State $[x_{0}^{img}, x_{0}^{txt}] = x_0  \sim p(x, c)$ \Comment{Sample image and text data}
    \State $t \sim \mathcal{U}(0,1)$ 
    \Comment{Sample random timestep}    
    \\      
    \State  $x_{t}^v \sim q(x_{t}^v \mid x_0^v) = \alpha_{t} x_0 + (1 - \alpha_{t})e_m \quad \text{for } v \in \{\text{img}, \text{txt}\}$ \Comment{mask all tokens}
    \\
    
    \State \textbf{With probability} $p_{\text{uncond}}$ \Comment{For Classifier-Free Guidance}:
    \State \quad \textbf{If} 
    $\text{rand}() < 0.5$:    \Comment{Randomly set one of the modalities to mask tokens}
    \State  \qquad $x_t^{img} \gets m$
    \State \quad \textbf{Else}: \\ \qquad \qquad $x_t^{txt} \gets m$

    \State  $x_0^{pred} = p_\theta([x_t^{img}, x_t^{txt}])$ \Comment{Estimate model prediction from masked sequence}
    \State Compute loss as:
    $
    \mathcal{L}_{\text{diff}} =
        \frac{\alpha'_t}{1 - \alpha_t} \log \langle x_0^{pred}, x_0 \rangle 
    $
    \Comment{Loss function over the logits of inputs}
    \State Perform gradient step on $\mathcal{L}$ to update $\theta$
\Until{converged}
\end{algorithmic}
\label{alg:training}
\end{algorithm}

\subsection{Sampling Algorithms}
\label{app:sampling_algo}
Here we describe the implementations of \model{}'s sampling algorithm and MaskGIT~\cite{chang2022maskgit}.

\ifthenelse{\boolean{arxiv}}
{}{
\begin{algorithm}[H]
\caption{\model{} Sampling}
\label{alg:sampling}

\begin{algorithmic}[1]
    \State \textbf{Initialize:} $x_T \gets [m, m, \ldots, m]$  \Comment{All tokens are masked}
    \State \textbf{Initialize:} $KV_{img} \gets \emptyset$  \Comment{Initialize KV-cache for image tokens to null set}
    \State \textbf{Require:} Sampling steps $T$, Text-image inference ratio $K$
    \State \textbf{Require:} Num Tokens to Unmask: $f(t)$. We set $f(t)$ as $\frac{1- \alpha_t}{\sum_{t=1}^{T} 1- \alpha_t}$
    \For{$t = T$ \textbf{down to} $1$}
        \State\textbf{If} $t$ modulo $K$ == 0: 
        \State \qquad $p_{x_0}, KV_{img} \gets p_\theta(x_0 \mid x_t)$ \Comment{Model predictions and  KV image tokens for caching. }
        \State\textbf{Else}: 
        \State \qquad $p_{x_0}^{txt} \gets p_\theta(x_0^{txt} \mid x_t^{txt}, KV_{img})$ \Comment{Use cached image tokens from previous timesteps}
        \State \qquad $p_{x_0} \gets  p_{x_0}^{txt}$ \Comment{Fix image tokens, Only sample text tokens}
        \State $p_{x_0}^{(k)} \gets \text{Top}_k(p_{x_0})$ \Comment{Top-$k$ filtering on logits}
        \State $p_{x_0}^{(k)} \gets \frac{p_{x_0}^{(k)}}{\tau(t)}$ \Comment{Apply temperature annealing}
        \State Sample $x_{\text{new}} \sim \text{Categorical}(p_{x_0}^{(k)})$ \Comment{Sample new tokens}
        \State $M \gets \left\lfloor f(t) \times N \right\rfloor$ \Comment{Determine number of tokens to unmask}
        \State Select $M$ most confident tokens based on $p_{x_0}^{(k)}$
        \State Update $x_{t-1}[i] \gets x_{\text{new}}[i] \quad \forall i \in \text{selected positions}$
        \State \textbf{Keep} previously unmasked tokens unchanged
    \EndFor
\end{algorithmic}
\end{algorithm}
}

\begin{algorithm}[H]
\caption{MaskGIT Sampling}
\begin{algorithmic}[1]
    \State \textbf{Initialize:} $x_T \gets [m, m, \ldots, m]$ \Comment{All tokens are masked}
    \State \textbf{Require:} Sampling steps $T$
    \State \textbf{Require:} Num Tokens to Unmask: $f(t)$. We set $f(t)$ as $\frac{1- \alpha_t}{\sum_{t=1}^{T} 1- \alpha_t}$
    \For{$t = T$ \textbf{down to} $1$}
        \State $p_{x_0} \gets p_\theta(x_0 \mid x_t)$ \Comment{Model prediction}
        \State $p_{x_0}^{(p)} \gets \text{Top}_p(p_{x_0})$ \Comment{Top-$p$ (Nucleus) sampling on logits}
        \State $p_{x_0}^{(k)} \gets \frac{p_{x_0}^{(k)}}{\tau(t)}$ \Comment{Apply temperature annealing}
        \State Sample $x_{\text{new}} \sim \text{Categorical}(p_{x_0}^{(k)})$ \Comment{Sample new tokens}
        \State $M \gets \left\lfloor f(t) \times N \right\rfloor$ \Comment{Determine number of tokens to unmask}
        \State Select $M$ most confident tokens based on $p_{x_0}^{(k)}$
        \State Update $x_{t-1}[i] \gets x_{\text{new}}[i] \quad \forall i \in \text{selected positions}$
        \State \textbf{Keep} previously unmasked tokens unchanged
    \EndFor
\end{algorithmic}
\end{algorithm}

\ifthenelse{\boolean{arxiv}}{}{
\newpage
\section{Extended Related Work}
\label{app:related_work}
\subsection{Unified Multi-Modal Models}
In recent years, unified models for processing multiple modalities have advanced significantly. Models like Flamingo~\cite{alayrac2022flamingo} and PaLM-E~\cite{Driess2023PaLMEAE} demonstrate strong few-shot learning capabilities across tasks. LLAVA~\cite{liu2023visualinstructiontuning} enhances LLaMa~\cite{touvron2023llama} with multimodal fine-tuning, but still uses separate encoders, limiting true unification and image generation. Recent efforts, like Perceiver IO~\cite{jaegle2021perceiver} and Unified-IO~\cite{lu2022unified}, attempt modality unification but at a smaller scale. The Chameleon project~\cite{chameleonteam2024chameleonmixedmodalearlyfusionfoundation} scales this up with a 34B parameter model trained on image-text data. However these approaches largely focus on autoregressive generation which is inefficient for high-dimensional data.

Relevant to our work, UniD3~\cite{hu2023unified} considered discrete diffusion on image and text but made several design decisions that separated each modality, using both absorbing and uniform masking, decoupling the modalities inside the model with separate operations on each. Further we couldn't compare against their model—no training code is available and were unable to reproduce their reported results using their publicly available code.

\subsection{Discrete Diffusion Models}
Discrete diffusion models have emerged as a promising alternative to continuous diffusion for discrete data types. \cite{JaschaSohl-DicksteinUnknown} introduced the first discrete diffusion model over binary variables,~\cite{HoogeboomUnknown} extended the noising process to categorical variables, demonstrating its effectiveness on image generation tasks. D3PM~\cite{Austin2021} later extended discrete diffusion to a more general set of noising processes, allowing for more flexible noise schedules. Recent work by SEDD~\cite{loudiscrete} introduced score entropy, a novel loss function for discrete diffusion models that bridges the gap between continuous and discrete spaces, and more recently, \cite{sahoo2024simple,shi2024simplified} showed text perplexity competitive with GPT-2. Most recently, \cite{nie2024scalingmaskeddiffusionmodels} looked at the scaling properties of discrete diffusion on text. While this approach shows promise for improving discrete diffusion models, these methods were primarily focused on language modeling tasks. Our work extends the application of discrete diffusion to multiple modalities and demonstrates its effectiveness in a unified architecture.
}

\newpage
\section{Additional Experiment Details}


\subsection{Conditional and Unconditional Experiment Details}
\label{app:cond_uncond_details}
For unconditional and conditional results in \cref{tab:uncond} and ~\ref{tab:cond} we use a dataset of 11B tokens comprising $30$M images from DataComp1B~\cite{gadre2024datacomp} and CC12M~\cite{changpinyo2021conceptual12mpushingwebscale} as our training set, with a fraction of $20\%$ text tokens and $80\%$ image tokens after excluding pad tokens. For faster convergence, we train only on DataComp1B for results in \cref{fig:inference} and \cref{fig:retrieve}. We tokenize the image and text tokens using separate tokenizers. We use  lookup-free quantization (LFQ) from~\cite{yu2023language,luo2024open} for as our image tokenizer, and use the tokenizer from~\cite{touvron2023llama} as our text tokenizer. We use an image resolution of $256\times256$, and a downsampling ratio of 16, resulting in a sequence length of $384$ with $256$ with image tokens and $128$ text tokens. Note that we use the same tokenizers for all the baselines, ensuring fair comparisons. We train \model{} for 300 L40S GPU hours and train the autoregressive model for a proportionate amount of time such that it achieves the same validation loss. Our model comprises 115M/340M non-embedding parameters and we use a batch size of $512$, a learning rate of $3\mathrm{e}{-4}$, and weight decay of $0.05$, following~\cite{sun2024autoregressive}.

\subsection{Conditional and Unconditional Evaluations}
\begin{figure*}[h!]
    \centering
    \includegraphics[width=0.7\columnwidth]{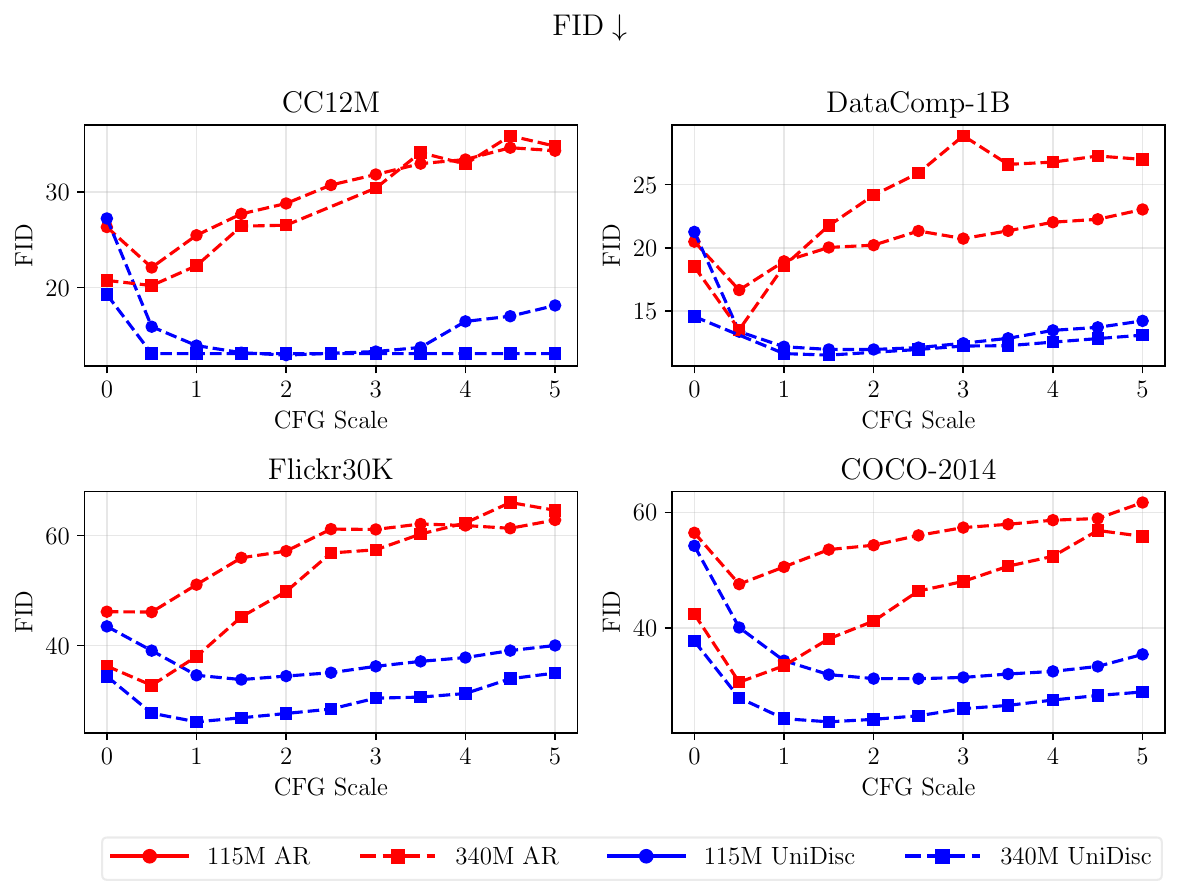}
    \includegraphics[width=0.7\columnwidth]{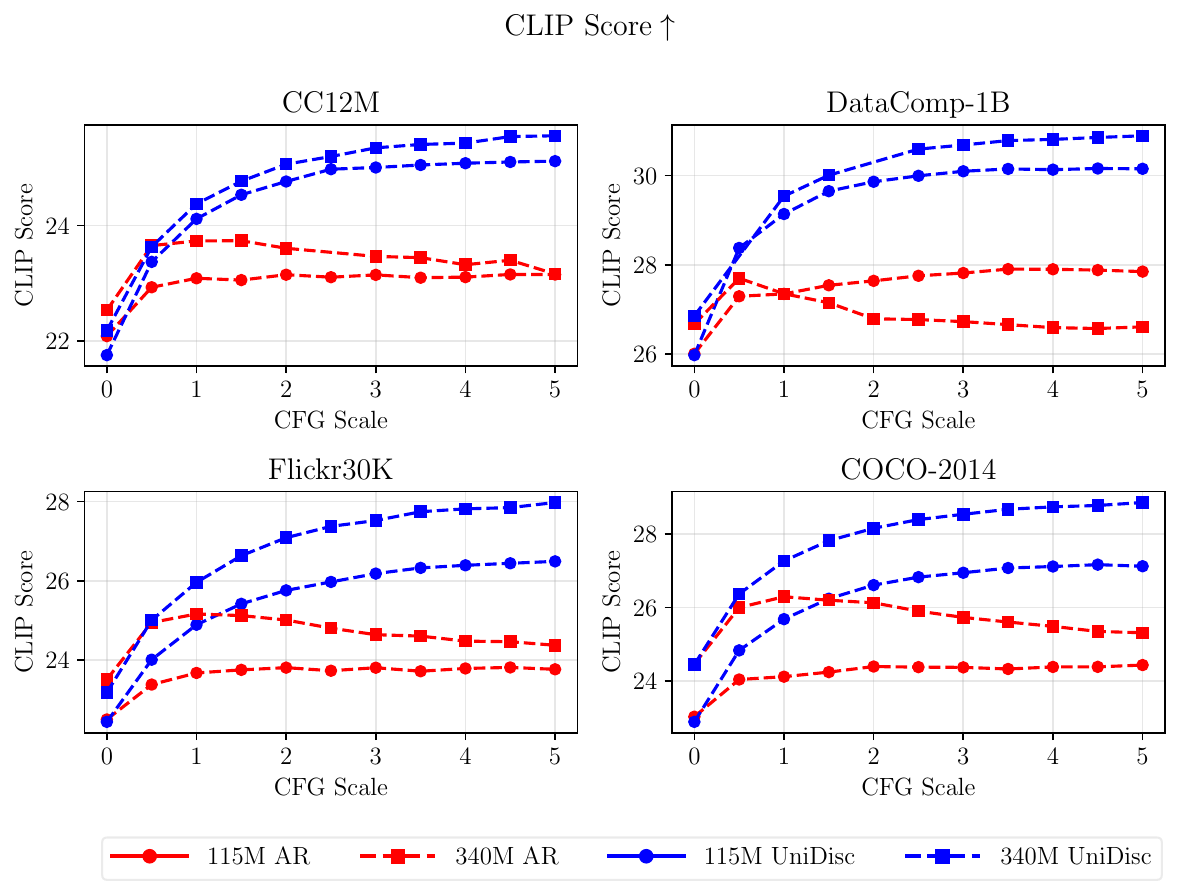}
    \vspace{-10pt}
    \caption{\textbf{Conditional generation results for both FID and CLIP metrics, across a range of CFG values.} We find that AR is more sensitive to the CFG weighting, with a narrower optimal range.}
    \label{fig:full_cfg_weight_ablation}
    \vspace{-10pt}
\end{figure*}

We extend \cref{fig:cfg_weight_ablation}, adding results on Flickr-30K and MS-COCO below in \cref{fig:full_cfg_weight_ablation}. We show unconditional results in \cref{tab:uncond} and conditional results (taking the optimal CFG weight for both \model{} and the AR model) in \cref{tab:cond}.

\begin{figure*}[h!]
    \centering
    \begin{minipage}{0.48\textwidth}
        \centering
        \renewcommand{\arraystretch}{1.5}
        \resizebox{\linewidth}{!}{
            \begin{tabu}{lccccccccccccccccccccc}
                & \rotatebox{0}{CC12M} & \rotatebox{0}{DataComp} & \rotatebox{0}{Flickr} & \rotatebox{0}{MS-COCO}\\
                \midrule
                \textbf{Image + Text Perplexity}  \\
                \hdashline
                Chameleon & 541.2 & 156.8 & 1254.9 & 1128.3   \\
                \hdashline
                \model{} & 494.5 & 154.8 & 1115.0 & 982.2  \\ 
                \midrule
                \textbf{Image - FID}  \\
                Chameleon & \textbf{30.5} & \textbf{20.49} & \textbf{75.70} & \textbf{70.67} \\
                \hdashline
                \model{} & 35.78 & 22.97 & 88.88 & 77.43  \\
                \midrule
                \textbf{Text - CLIP}  \\
                Chameleon & 23.70 & \textbf{26.08} & 23.70 & 23.64  \\
                \hdashline
                \model{} & \textbf{25.01} & 25.98 & \textbf{24.92} & \textbf{25.01}  \\      
                \bottomrule
            \end{tabu}
        }
        \caption{Unconditional multimodal generation results for \model{} and AR baseline at 115M parameters - both models perform similarly.}
        \label{tab:uncond}
    \end{minipage}
    \hfill
    \begin{minipage}{0.48\textwidth}
        \centering
        \renewcommand{\arraystretch}{1.5}
        \resizebox{\linewidth}{!}{
            \begin{tabu}{lcccc}
                & \rotatebox{0}{CC12M} & \rotatebox{0}{DataComp} & \rotatebox{0}{Flickr} & \rotatebox{0}{COCO} \\
                \midrule
                \textbf{Text to Image - FID}  \\
                Chameleon 115M w/o CFG & 26.32 & 20.49 & 46.13 & 56.46 \\
                Chameleon 340M w/o CFG & 20.75 & 18.53 & 36.24 & 42.41 \\
                Chameleon 115M w/ CFG (0.5) & 22.10 & 16.68 & 46.06 & 47.58 \\
                Chameleon 340M w/ CFG (0.5) & 20.22 & 13.55 & 32.74 & 30.62 \\
                \hdashline
                \model{ 115M} w/o CFG & 27.22 & 21.26 & 43.46 & 54.21 \\
                \model{ 340M} w/o CFG & 19.28 & 14.59 & 34.37 & 37.73 \\
                \model{ 115M} w/ CFG (1.5) & \textbf{13.21} & \textbf{12.00} & \textbf{33.79} & \textbf{31.94} \\
                \model{ 340M} w/ CFG (1.5) & \textbf{13.11} & \textbf{11.55} & \textbf{26.83} & \textbf{23.77} \\
                \midrule
                \textbf{Image to Text - CLIP}  \\
                Chameleon 115M w/o CFG & 22.08 & 26.01 & 22.50 & 23.02 \\
                Chameleon 340M w/o CFG & 22.53 & 26.68 & 23.51 & 24.46 \\
                Chameleon 115M w/ CFG (0.5) & 22.93 & 27.30 & 23.38 & 24.03 \\
                Chameleon 340M w/ CFG (0.5) & 23.65 & 27.70 & 24.95 & 25.99 \\
                \hdashline
                \model{ 115M} w/o CFG & 21.75 & 25.98 & 22.44 & 22.88 \\
                \model{ 340M} w/o CFG & 22.18 & 26.86 & 23.18 & 24.44 \\
                \model{ 115M} w/ CFG (1.5) & \textbf{24.54} & \textbf{29.65} & \textbf{25.42} & \textbf{26.24} \\
                \model{ 340M} w/ CFG (1.5) & \textbf{24.77} & \textbf{30.01} & \textbf{26.63} & \textbf{27.82} \\
                \bottomrule
            \end{tabu}
        }
        \caption{Conditional generation results for \model{} and AR baseline. Our model significantly outperforms the AR model when classifier free guidance is used.}
        \label{tab:cond}
    \end{minipage}
\end{figure*}

\subsection{Generative Perplexity — Qualitative}
\label{app:low_pplx_examples}
\begin{table}[h]
\centering
\scriptsize 
\begin{tabular}{|p{4.2cm}|c|c|}
\hline
\textbf{Text} & \textbf{Chameleon Perplexity} & \textbf{GPT2 Perplexity} \\ \hline
"ICLR is globally renowned for presenting..." (Continued) & 32.836 & 35.780 \\ \hline
"This is simple. This is simple." (Repeated) & 8.423 & 3.930 \\ \hline
"Words Words Words Words" (Repeated) & 2.226 & 3.583 \\ \hline
"AAAAAAAAAAAA" (Repeated) & 2.732 & 1.904 \\ \hline
"(Spaces Repeated)" & 80.240 & 1.095 \\ \hline
\end{tabular}
\caption{We demonstrate how generative perplexity is an imperfect metric requiring calibration with entropy.}
\label{tab:perplexity}
\end{table}

\newpage
\subsection{Quantitative Inpainting Comparison w/autoregressive models}
\label{ssec:inpainting_ar_nar}
\begin{figure}[H]
    \centering
    \includegraphics[width=1.0\linewidth]{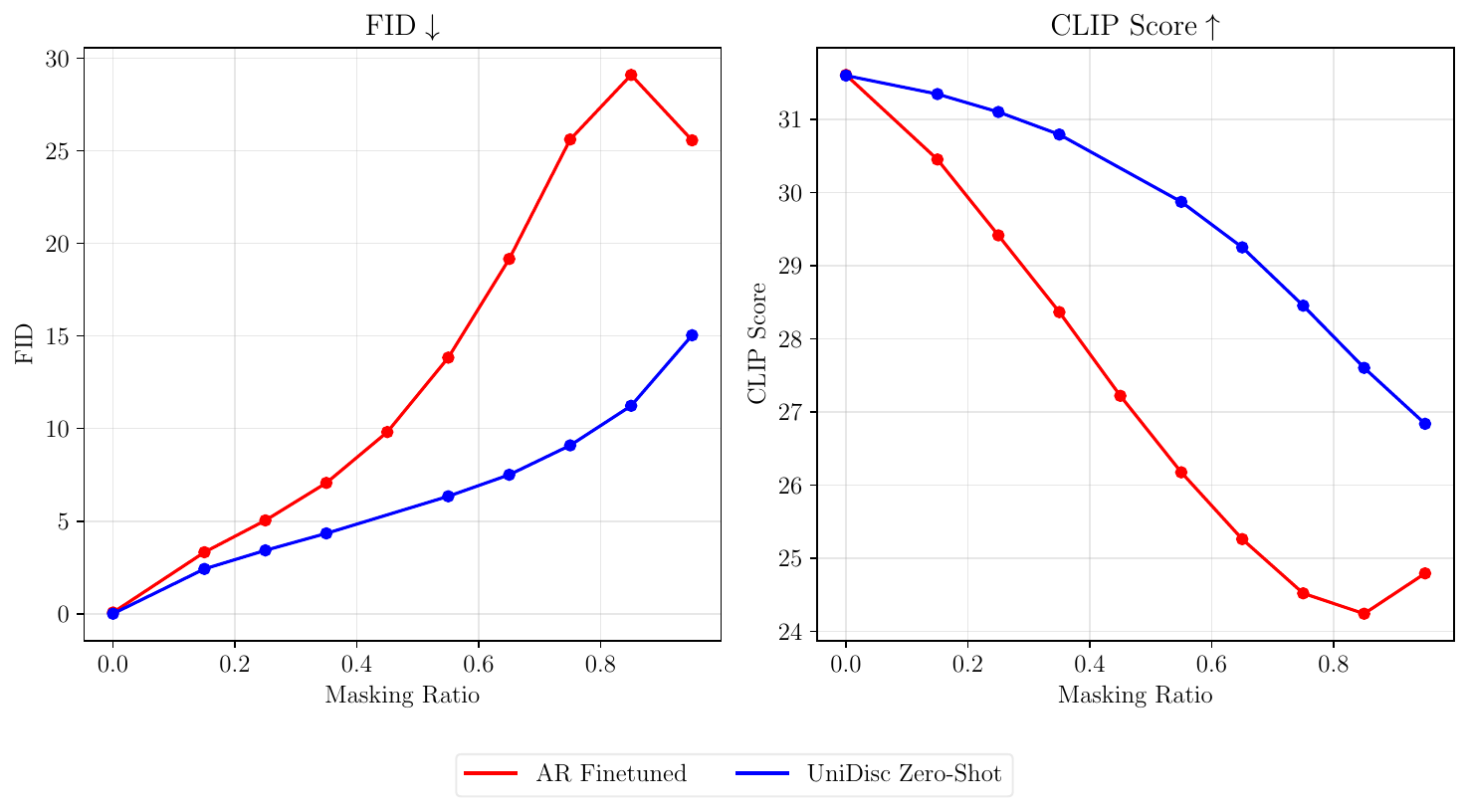}
    \caption{We compare \model{} with an AR model fine-tuned for joint inpainting and evaluate on a subset of DataComp1B.}
    \label{fig:inpainting_comparison}
\end{figure}

To demonstrate the tradeoff between the pre-training objectives of \model{} and AR models, we evaluate both models on inpainting. We fine-tune the 340M parameter AR model on a standard set of multimodal datasets (CC12M, Recap-DataComp-1B, LAION 400M) and evaluate \model{} in a zero shot manner—without any fine-tuning. Specifically, for the AR model, we use a linear masking schedule for the prefix sequence consisting of a randomly masked text and image pair and then predict and supervise the clean sequence, doubling the overall sequence length. In \cref{fig:inpainting_comparison}, we evaluate at multiple noise levels, showing the degradation in performance as the original sequence is increasingly masked.

\subsection{Discriminative Evaluations}
\label{ssec:discrim_evaluations}

For evaluations on CLEVR-VQA and CLEVR-Ref ~\cite{liu2019clevrrefdiagnosingvisualreasoning} we use their respective train-val splits. Note that for CLEVR-VQA and CLEVR-Ref, we do not follow the training scaling factor found in \cref{fig:scaling_laws}, we instead train both the models until convergence, i.e multiple epochs. The small size of these datasets makes it possible to train until convergence. For CLEVR images, we find that none of the existing tokenizers work well, so we fine-tune our own tokenizer on CLEVR images. We use images of $128\times128$ resolution, with a total sequence length of 320 (256 image tokens  and 64 text tokens). For text, we use a standard BERT tokenizer~\cite{devlin2019bertpretrainingdeepbidirectional}. In Figure \cref{fig:retrieve}, we ablate the role CFG and the number of denoising steps play in \model{}'s retrieval accuracy. While the number of denoising steps in an AR model is fixed to the sequence length, the denoising steps for \model{} can be much higher.
\begin{figure}[h!]
    \centering
    \includegraphics[width=0.6\columnwidth]{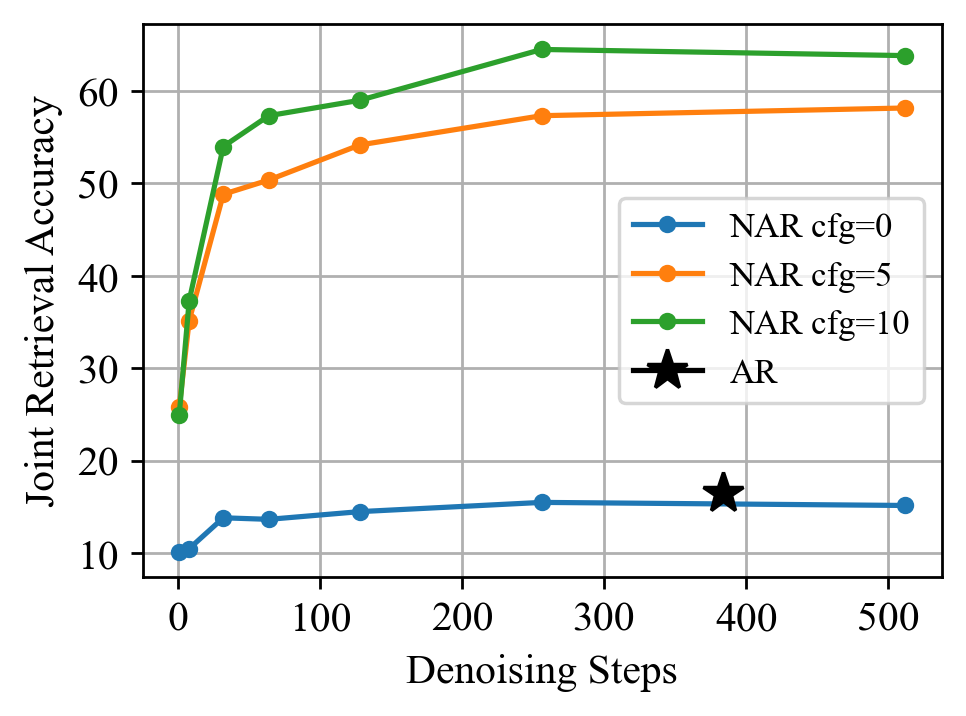}
    \vspace{-10pt}
    \caption{\textbf{Joint Retrieval Accuracy on DataComp1B.} We outperform AR given the task of retrieving one correct image-text pair out of 16 possible pairs, implying better learnt representations.}
    \label{fig:retrieve}
    \vspace{-2pt}
\end{figure}

\newpage
\section{Scaling Experiment Details}
\label{app:scaling}
As in prior experiments, all implementation details are shared between \model and AR training configurations except: (1) causal vs. full attention, (2) masking of the input sequence for \model, and (3) weighting the CE loss as in \cref{eq:mdlm}.

We use similar hyperparameters as in our small-scale experiments in \cref{exp_details}, but repeat them here for clarity. For images, we use lookup-free quantization (LFQ)~\cite{yu2023language,luo2024open} with a image resolution of $256\times256$, and a downsampling ratio of 16, resulting in a sequence length of $256$ image tokens. We use a BPE tokenizer~\cite{touvron2023llama} for text with $128$ text tokens, resulting in a total sequence length of $384$. We report only non-embedding parameters and data tokenization is identical across all models.

We use a batch size of $512$ and use AdamW~\cite{loshchilovDecoupledWeightDecay2019} with $\beta_1=0.9$, $\beta_2=0.95$, and weight decay $\lambda=0.05$. We use a max learning rate of $3\mathrm{e}{-4}$ with a linear warmup followed by a cosine decay schedule, ending with zero at the final step for a given training run. 


We list all model variants in \cref{tab:model_variants}.

\begin{table}[h]
    \centering
    \begin{tabular}{r|r|r|r}
        \toprule
        Parameters (M) & n\_layers & n\_heads & d\_model \\
        \midrule
        34  & 11  & 6  & 384  \\
        67  & 11  & 9  & 576  \\
        116 & 12  & 12 & 768  \\
        172 & 20  & 12 & 768  \\
        228 & 20  & 14 & 896  \\
        343 & 24  & 16 & 1024 \\
        484 & 22  & 10 & 1280 \\
        543 & 17  & 12 & 1536 \\
        622 & 29  & 10 & 1280 \\
        713 & 23  & 12 & 1536 \\
        826 & 27  & 12 & 1536 \\
        1074 & 26  & 14 & 1792 \\
        1290 & 24  & 16 & 2048 \\
        \bottomrule
    \end{tabular}
    \caption{Model variants. The FFN hidden size is always 4x the overall $\operatorname{d\_model}$}
    \label{tab:model_variants}
\end{table}

\newpage
\section{Training Details}
\subsection{Additional Training Implementation Details}
We use flash attention for all models except as noted below, using the popular Flash-Attention 2 library~\cite{daoFlashAttention2FasterAttention2023}. For all AR models at inference, we use K/V caching and take advantage of specially optimized functions for this in FlashAttention 2.


\section{Fine-tuning An Autoregressive model for Discrete Diffusion}
As we already have a plethora of large-scale AR models~\cite{touvron2023llama,chameleonteam2024chameleonmixedmodalearlyfusionfoundation}, it would be useful to have the ability to fine-tune them for a discrete diffusion objective. While the naive method for fine-tuning would be to change the objective function to discrete diffusion while using AR's pre-trained weights. We find that a better idea is to left-shift the output targets of the diffusion objective such that instead of having the masked token predict its respective visible token, we have the token before the masked token predict it. In this way, we more closely match the original AR next-token prediction objective. In \cref{fig:finetune_model} we show that this strategy works well and we can effectively fine-tune a pre-trained autoregressive language model using discrete diffusion loss. We demonstrate this result on a 270M parameter language model~\cite{mehta2024openelm}, OpenELM, which is trained with an AR objective. We compare against training from scratch and training AR without the shift. We find the shifting strategy converges faster.

\subsection{Large Scaling Training Curve}
\label{app:large_scale_curve}
\begin{figure*}[h!]
    \centering
    \begin{minipage}[t]{0.48\linewidth}
        \centering
        \includegraphics[width=\linewidth]{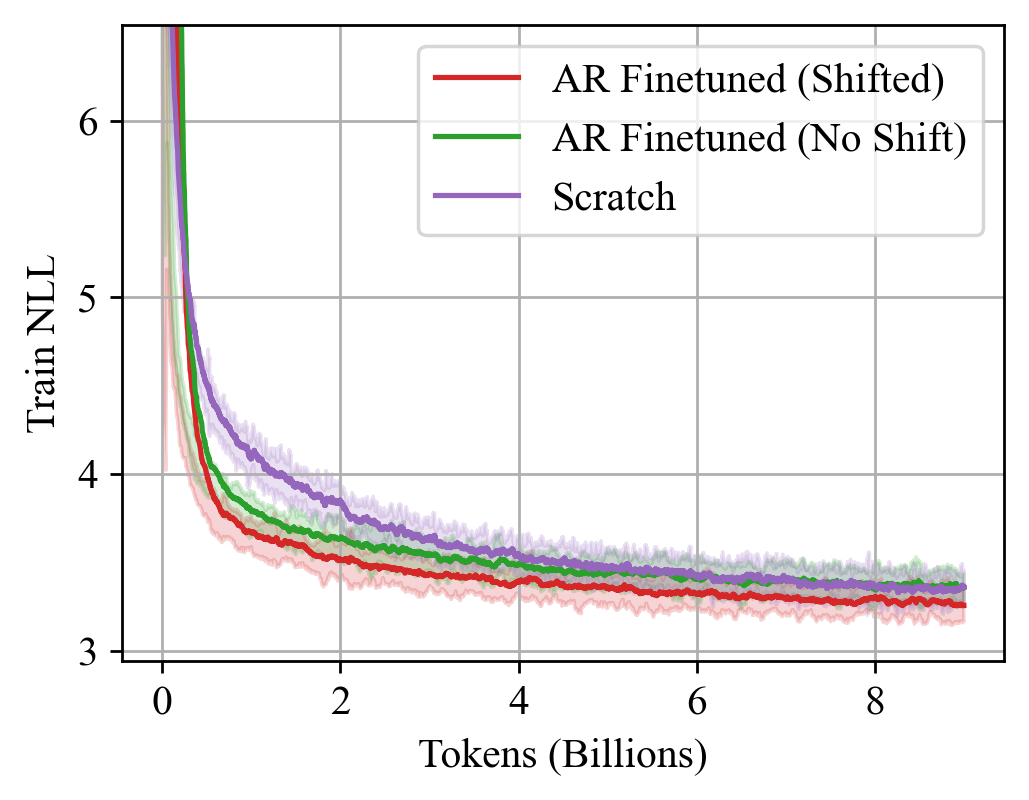}
        \caption{Fine-tuning a pre-trained 270M parameter AR model on LM1B.}
        \label{fig:finetune_model}
    \end{minipage}
    \hfill
    \begin{minipage}[t]{0.48\linewidth}
        \centering
        \includegraphics[width=\linewidth]{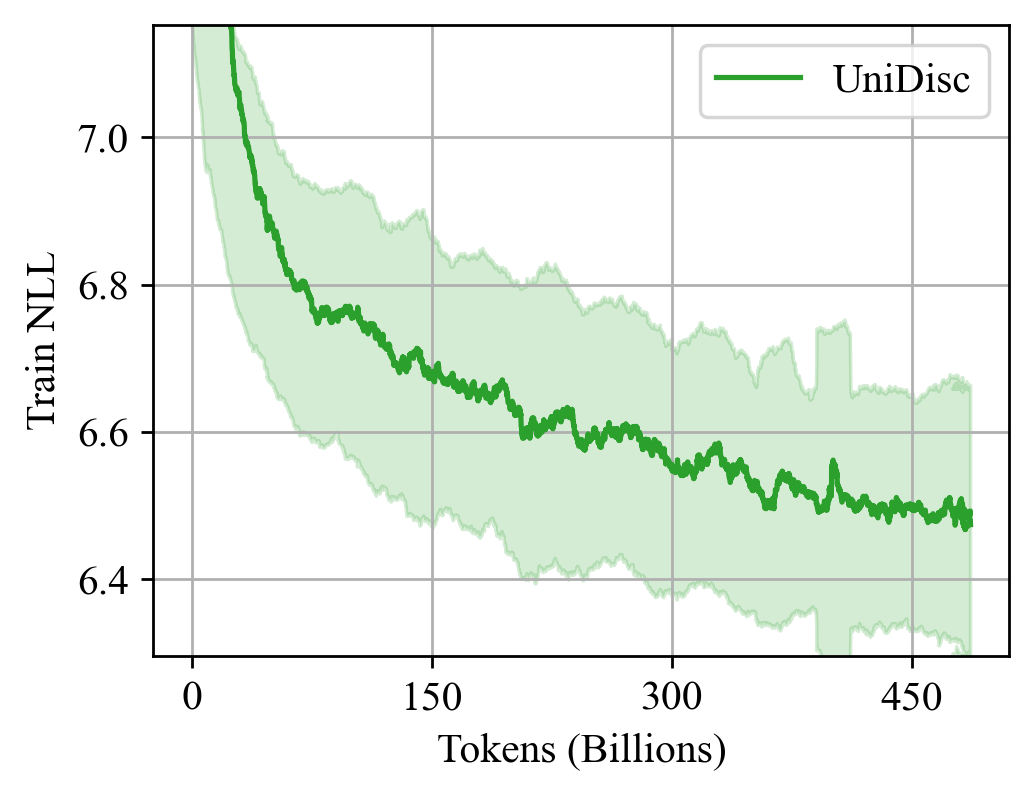}
        \caption{Training Loss Curve vs. Tokens on our 1.4B model.}
        \label{fig:finetune_large_scale_curve}
    \end{minipage}
\end{figure*}

We show the training curve for the large scale experiments described in \cref{sec:large_scale} in \cref{fig:finetune_large_scale_curve}.

\newpage
\section{Ablations}
\label{app:ablations}
We validate our design choices by running small-scale experiments on a subset of our primary dataset, taking 18M image/caption pairs on DataComp1B. We train on lower-resolution images at $128\times128$ and obtain a $1$:$1$ ratio of text to image tokens, with $64$ text and $64$ image tokens for a total sequence length of $128$, with all other hyperparameters the same as in our primary experiments.

We examine the influence of several design choices for our model in Table~\ref{tab:supervised_cls} and reach several conclusions. First, architecture changes to improve training stability—namely adding QK Normalization and using RMSNorm instead of LayerNorm—do not substantially affect convergence in this setting.

Another natural design choice is to parameterize the model such that we provide the modality of a given token to the model. With this relaxation we can drastically reduce the output space and, in theory, simplify the objective for our model. However, we find that this reparametrization only marginally reduces overall perplexity, even at this smaller-scale. We hypothesize that the modality-specific embeddings added to each token allows the model to learn the correct output space with minimal added parameters.

\begin{table}[H]
\begin{minipage}[t]{0.48\textwidth}
    \centering
    \renewcommand{\arraystretch}{1.5}
    \resizebox{0.95\linewidth}{!}{
    \begin{tabular}{lcccc}
        \toprule
        & \textbf{DataComp1B Validation PPL}  \\
        \midrule
        \model{} & 93.8 \\
        \hspace{0.5cm}w/o QK Norm & 92.7 \\
        \hspace{0.5cm}w/ Zero-linear init & 93.8 \\
        \hspace{0.5cm}w/o RMSNorm & 93.8 \\
        \hspace{0.5cm}w/o -inf for invalid tokens & 94.7 \\
        \hspace{0.5cm}w/o Softmin SNR & 109.6 \\ 
        \hspace{0.5cm}None & 111.2 \\ 
        \bottomrule
    \end{tabular}
    }
    \caption{Ablation w/115M parameter model of QK Norm, zero initialization of linear layers, RMSNorm, setting invalid tokens to $-\infty$ during training and generation, and Softmin SNR.}
    \label{tab:supervised_cls}
\end{minipage}
\hfill
\begin{minipage}[t]{0.48\textwidth}
    \centering
    \renewcommand{\arraystretch}{1.5}
    \resizebox{0.95\linewidth}{!}{
    \begin{tabular}{lcccc}
        \toprule
        & \textbf{DataComp1B Validation FID}  \\
        \midrule
        \model{} & 11.4 \\
        \hspace{0.5cm}w/cosine noising schedule & 11.5 \\
        \hspace{0.5cm}w/o CE loss weighting & 11.35 \\ 
        \hspace{0.5cm}w/discrete time (T=1000) & 13.8 \\ 
        \bottomrule
    \end{tabular}
    }
    \caption{Ablation w/115M parameter model on different objective level decisions such as noising schedule, loss weighting and whether to use discrete time.}
    \label{tab:objective_ablation}
\end{minipage}
\end{table}

\newpage
\section{Large Scale Qualitative Results}
\label{app:large_scale_extra_qual}
\begin{figure}[H]
    \centering
    \includegraphics[width=1.0\linewidth]{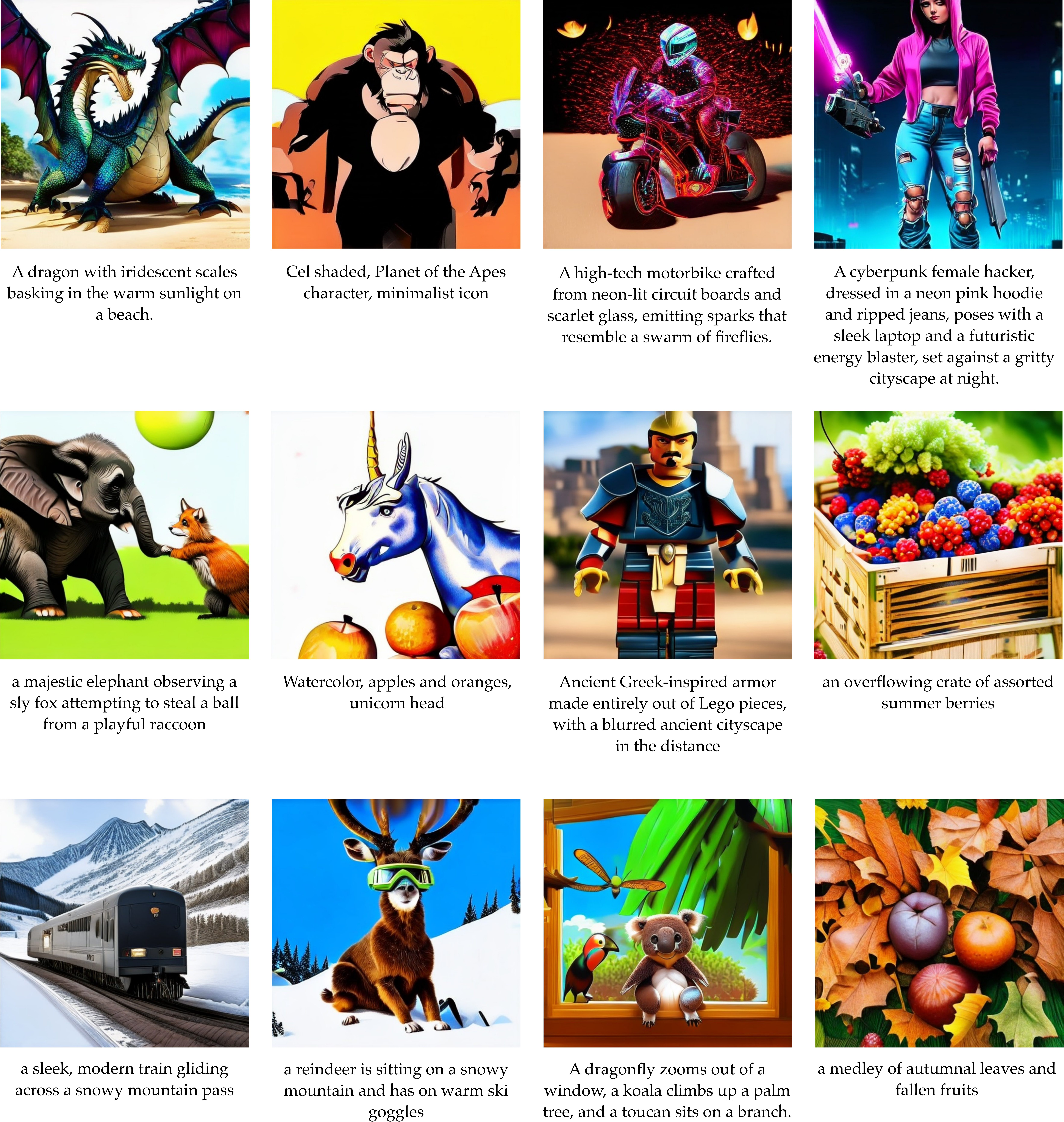}
    \caption{\model{}'s ability to generate an image, given unseen text as input.}
    \label{fig:t2i}
\end{figure}

\begin{figure}[H]
    \centering
    \includegraphics[width=1.0\linewidth]{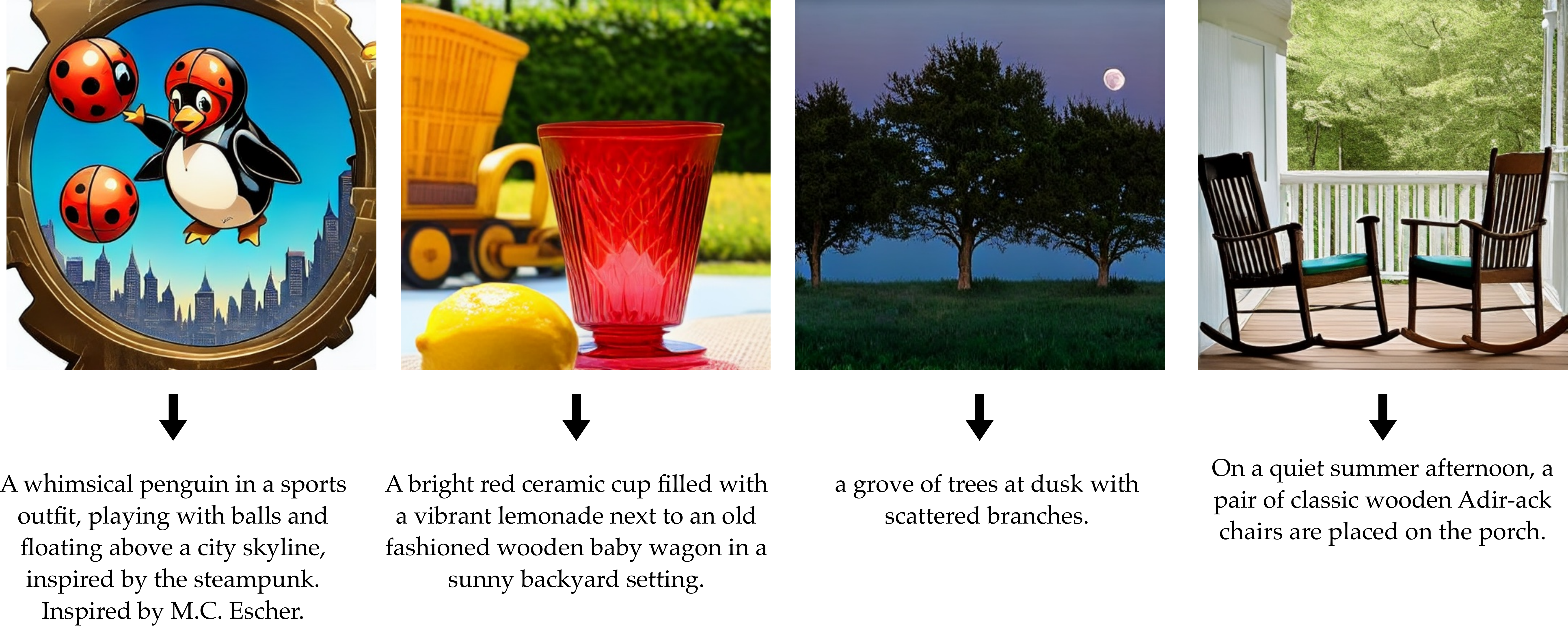}
    \caption{\model{}'s ability to generate text (captioning), given unseen image as input.}
    \label{fig:i2t}
\end{figure}

\begin{figure}[H]
    \centering
    \includegraphics[width=1.0\linewidth]{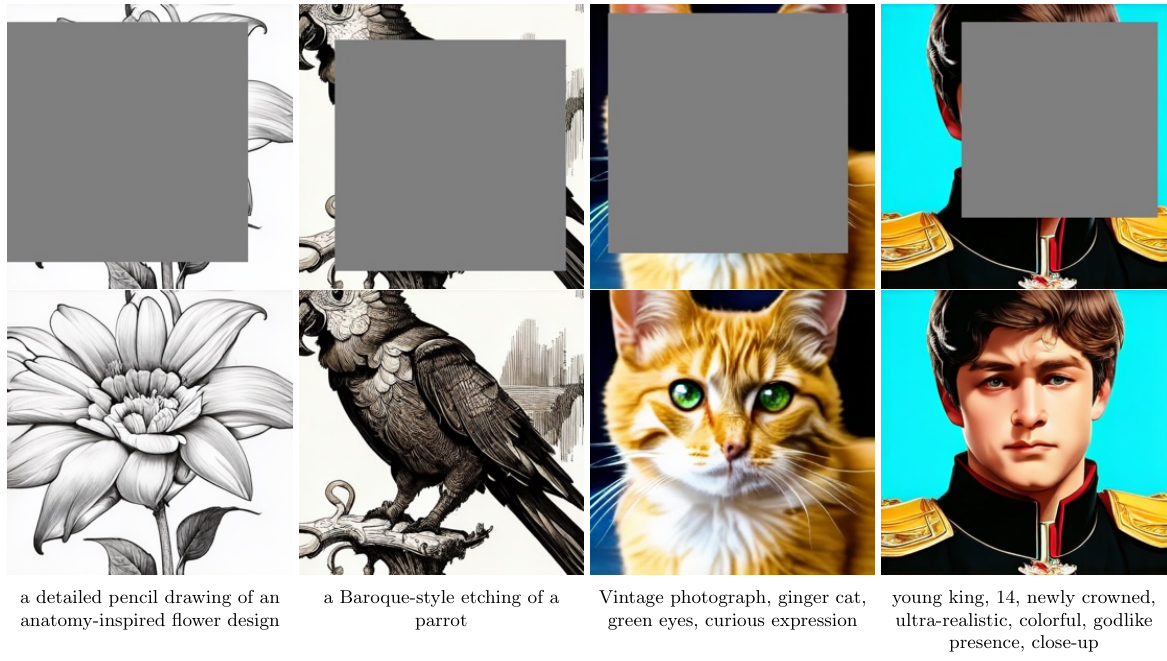}
    \caption{Zero-shot text-conditioned inpainting. \model{} inpaints a masked region given a user-provided text prompt.}
    \label{fig:text_cond_image_inpaint}
\end{figure}

\begin{figure}[H]
    \centering
    \includegraphics[width=1.0\linewidth]{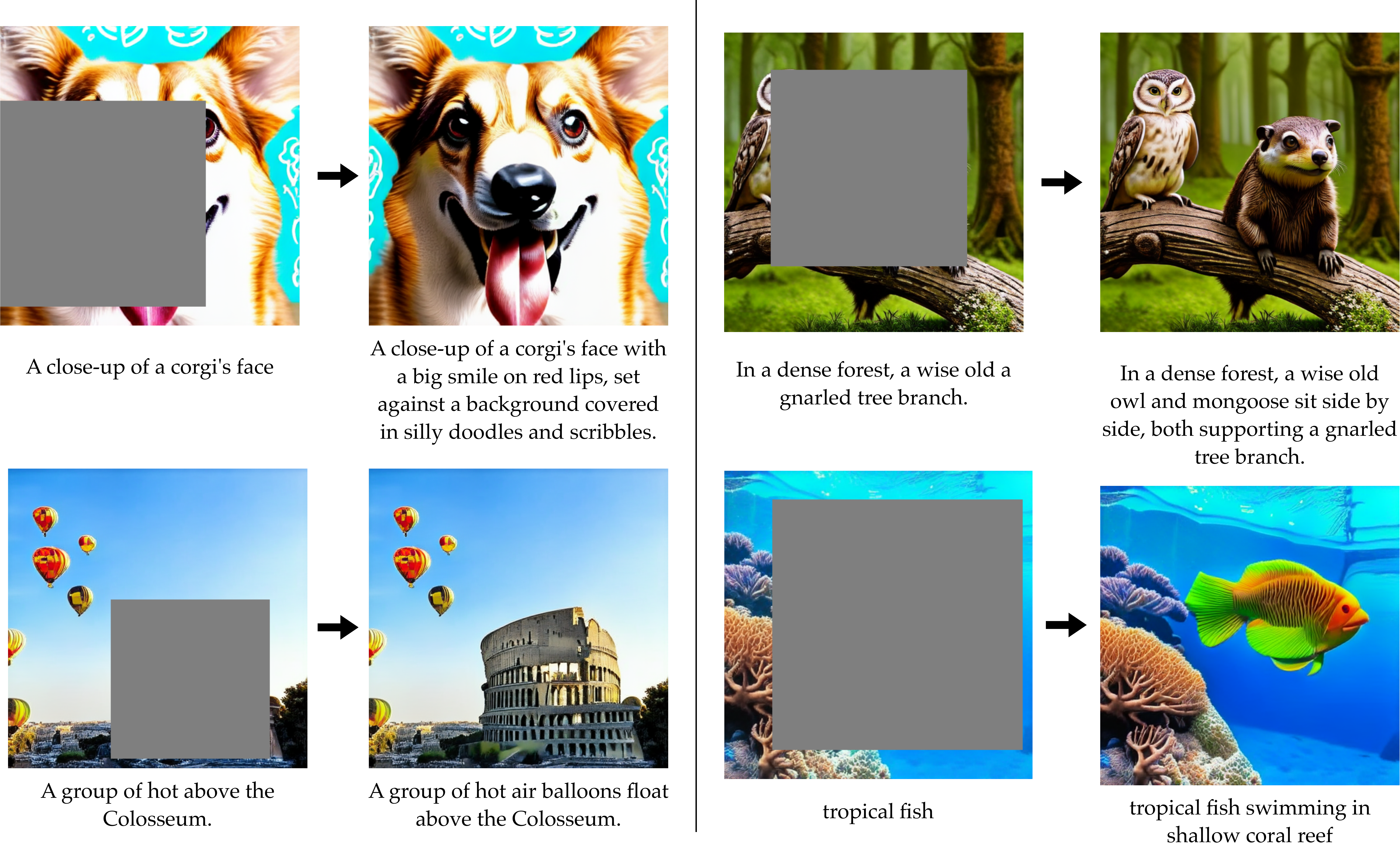}
    \caption{Zero-shot multimodal inpainting. \model{} jointly inpaints in both image and text spaces.}
    \label{fig:appendix_joint_image_inpaint}
\end{figure}

We show additional results on tasks such as joint inpainting, image captioning and image generation. We note that none of these tasks were explicitly trained or optimized for by our model. This is an intrinsic property due to the nature of \model{}'s unified diffusion based objective. In \cref{fig:t2i} we show standard text-to-image generation and in \cref{fig:i2t} we show standard image-to-text generation. In \cref{fig:text_cond_image_inpaint} we show zero-shot text-conditioned inpainting, and in \cref{fig:appendix_joint_image_inpaint} we show zero-shot \textit{multimodal} inpainting.

\newpage
\subsection{Zero-shot image editing of \model{}}
\label{ssec:automatic_image_editing}
\begin{figure}[H]
    \centering
    \includegraphics[width=0.85\linewidth]{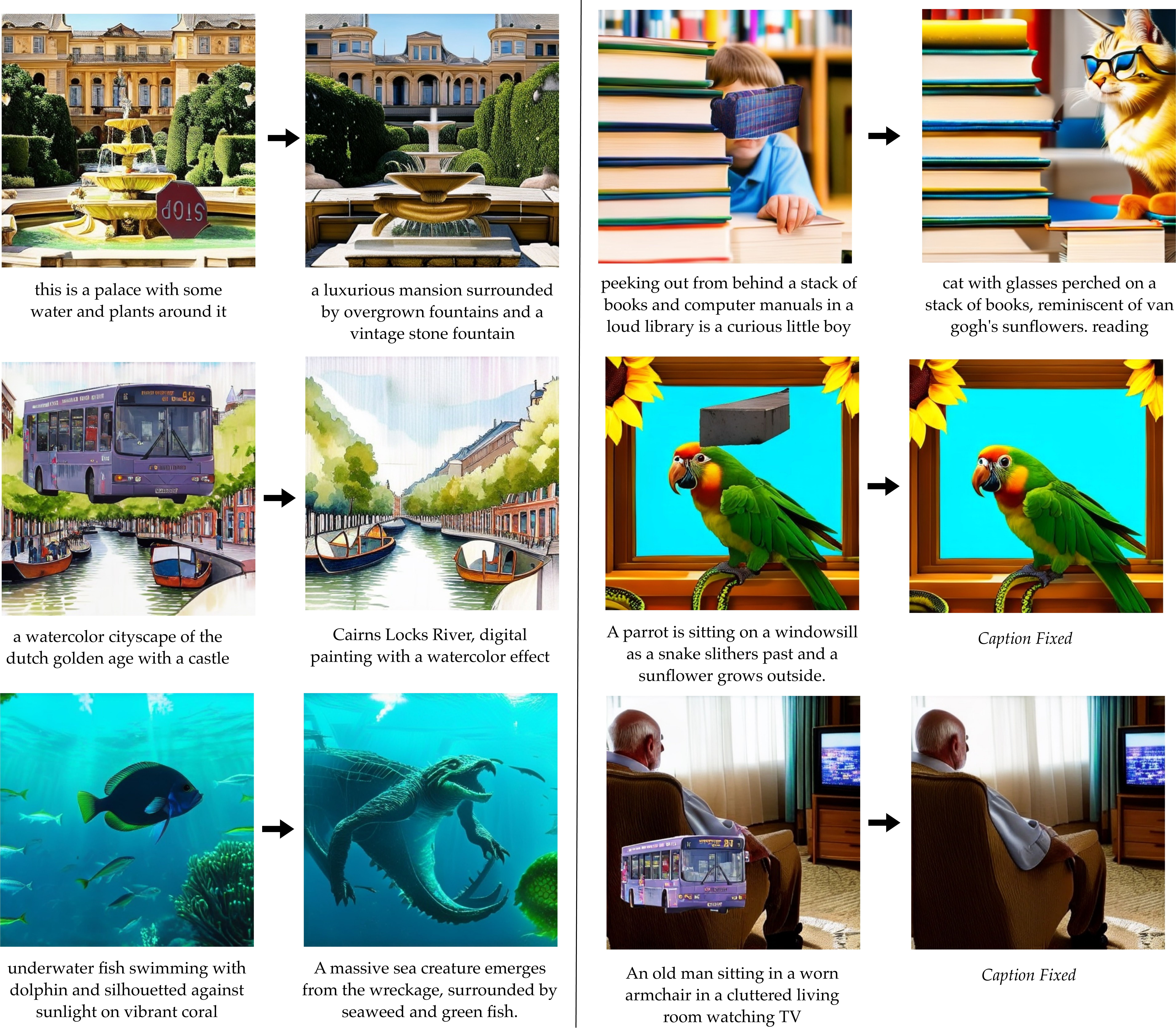}
    \caption{Zero-shot multimodal editing. We provide a \textit{clean} image and text pair and \model{} automatically enhances both the image and text. In the final row, we fix the text and allow only the image to change.}
    \label{automaticenhancement}
\end{figure}

A clear benefit of diffusion models is the ability to perform zero-shot editing without specific paired data—which is often difficult to obtain. We demonstrate one such method in Figure~\ref{automaticenhancement}, showing that \model{} can automatically improve a user provided image and caption.

We augment real images by overlaying random objects from the COCO dataset. Similarly, we augment captions by asking an LLM to generate purposely incorrect variations. We then randomly mask the image and text inputs and unmask as described above, automatically removing these undesired image artifacts and generating the correct caption. We adopt a best-of-n sampling strategy with n distinct noise masks. We unroll each generation using the model's own likelihood to select the best generation.

\vspace{-10pt}

\subsection{Analyzing the joint image-text generation of \model{}}
\label{ssec:joint_generation_analysis}
\begin{figure}[H]
    \centering
    \includegraphics[width=0.95\linewidth]{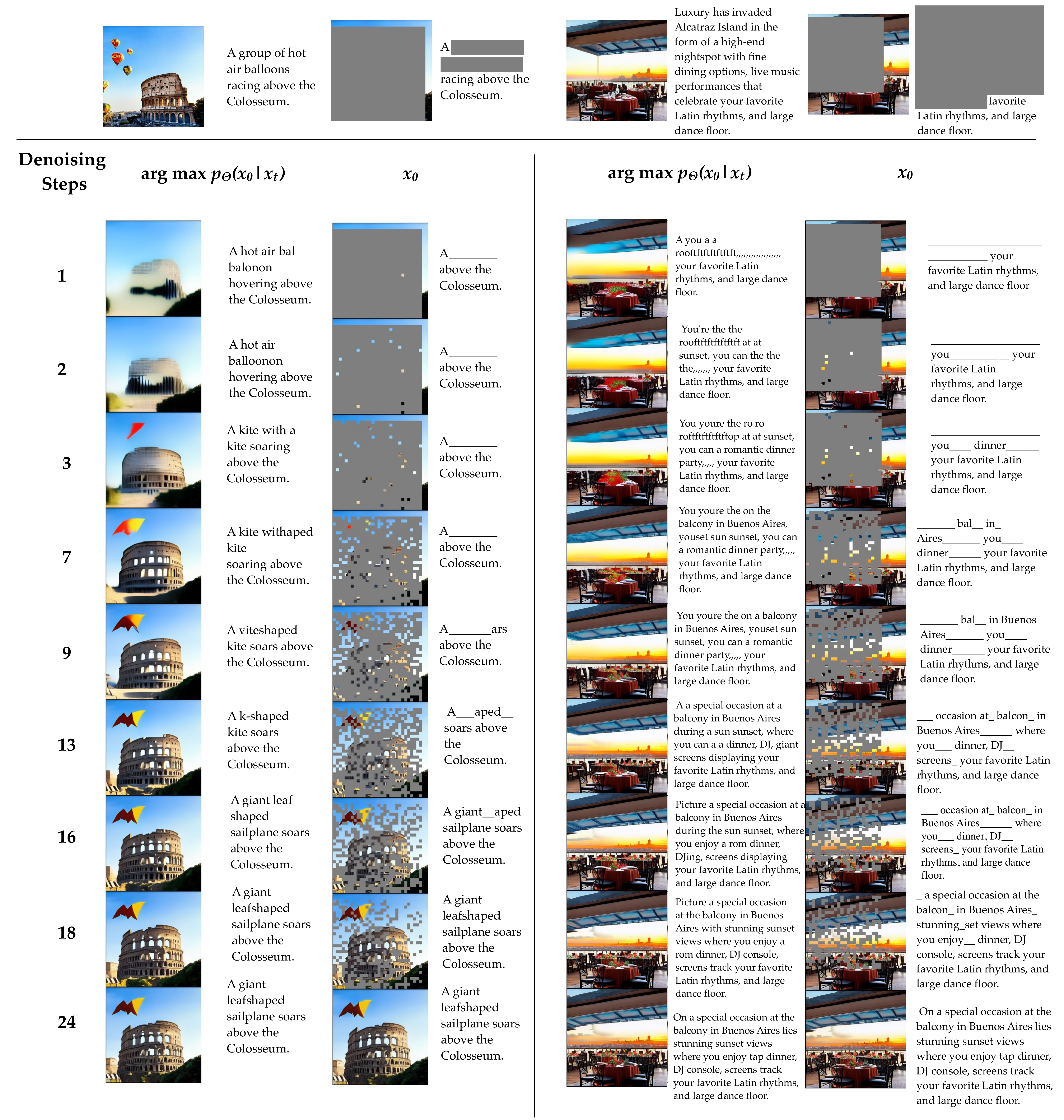}
    \caption{We show how \model{} jointly infills both image and text. $\operatorname{argmax} p_\theta (x_0 \mid x_t)$}
    \label{fig:intermediate_viz}
\end{figure}

\begin{figure}[H]
    \centering
     \includegraphics[width=0.80\linewidth]{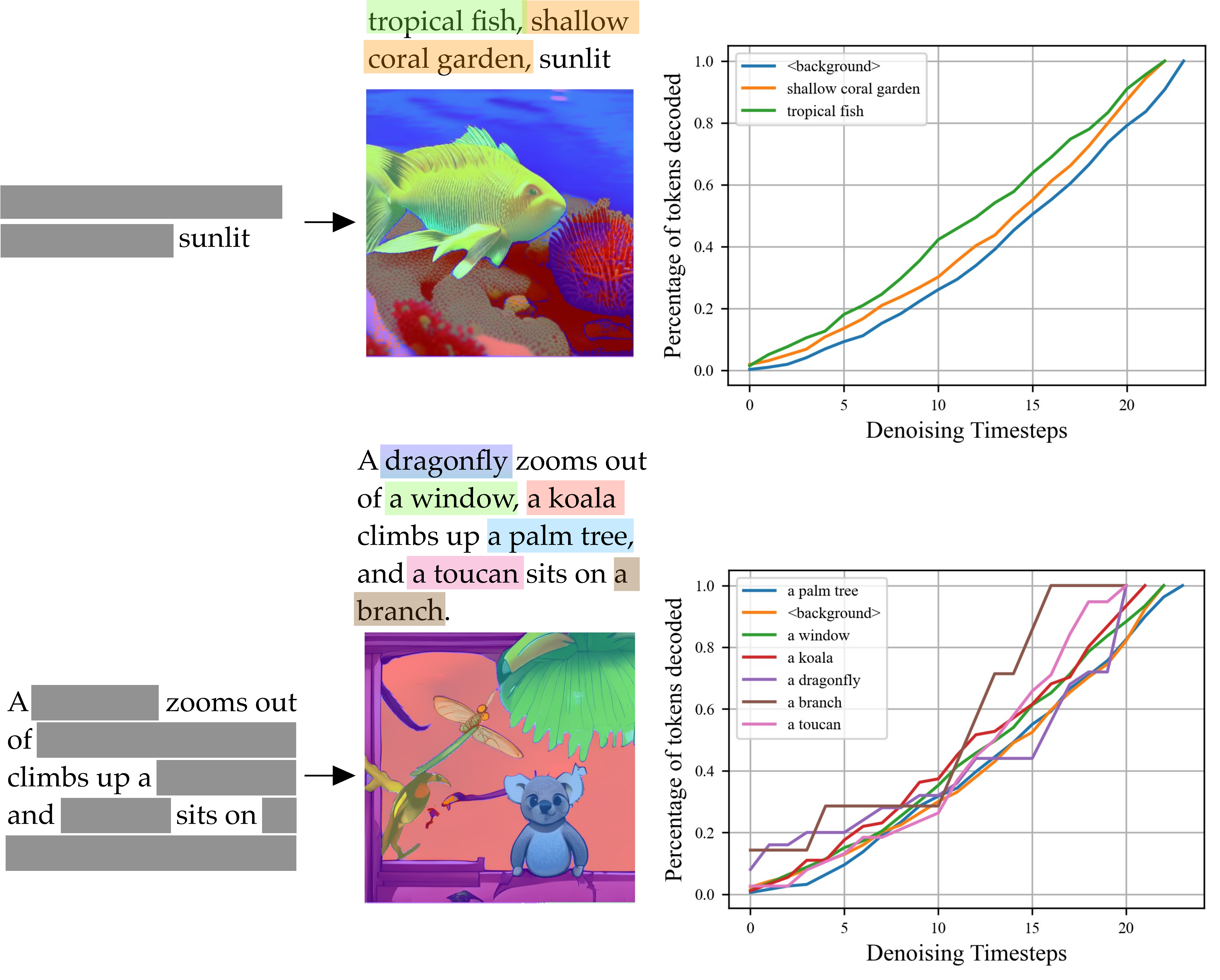}
    \caption{We demonstrate that \model{} uniformly generates all concepts at once.}
    \label{fig:segmentation_int}
\end{figure}

In Figure \ref{fig:intermediate_viz}, we visualize how the model iteratively infills both image and text. This raises the question - does \model{} follow a certain strategy during generation (for example, generating entire background first then moving to subject or generating text first before image), or does it generate everything at once jointly. To analyze this, we take the final model generated image, semantically segment it (using Grounded SAM 2 in our case) and then see which concepts get generated at what timesteps. This is visualized in Figure \ref{fig:segmentation_int}. We find that \model{} generates all concepts at once proportional to the overall fraction of the image the concept occupies. We also investigated if the \model{} has any strong positional bias, such as first generating tokens in the middle and radially filling out. However we find no such positional strategy and that \model{} is positionally invariant. Intuitively, this means that at any denoising step, all positions are equally likely to be decoded.

\subsection{Zero-shot length extrapolation of \model{}}
\begin{figure} [H]
    \centering
    \includegraphics[width=1.0\linewidth]{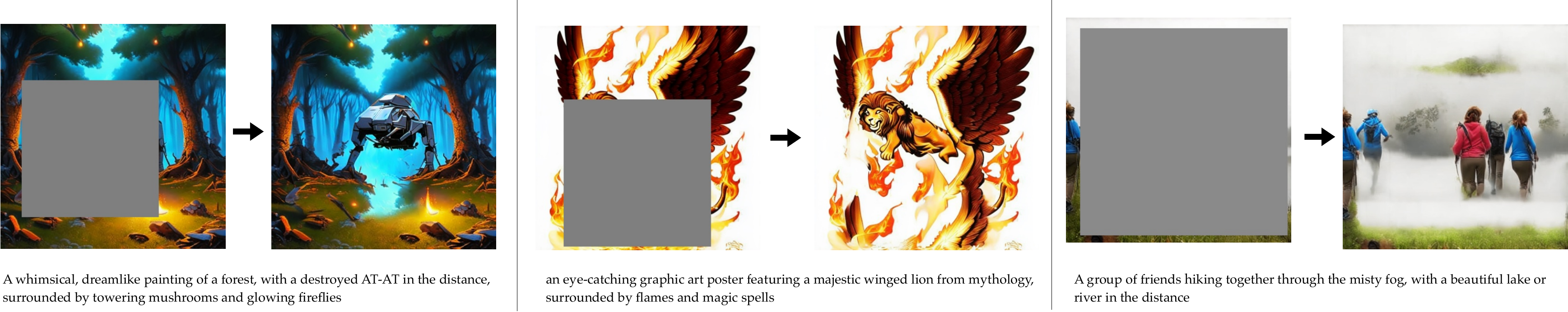}
    \caption{We train \model{} on 512x512 resolution images but demonstrate zero-shot inpainting at 1024x1024.}
    \label{extrapolation}
\end{figure}

In this section, we demonstrate the ability of \model{} to perform zero-shot flexible resolution generation thanks to the use of RoPE embeddings on both text and image tokens. \model{} model was fine-tuned on 512x512 images—resulting in each image using 1024 tokens—but is able to infill at 1024x1024—resulting in 4096 tokens per image—without further training. We demonstrate this in \cref{extrapolation}.

\newpage
\section{Large Scale Quantitative Comparisons}
\label{sec:large_scale_quantitative}
\subsection{Quantitative Generation Comparison with recent mulitmodal models}
In \cref{tab:geneval}, we evaluate \model{} on the popular GenEval~\cite{ghosh2023genevalobjectfocusedframeworkevaluating} benchmark which looks at how well a generated image adheres to the prompt in terms of a set of predefined attributes (e.g., color, positioning). In \cref{tab:fid_scores_large}, we compare FID on the popular MS-COCO 30K~\cite{chen2015microsoftcococaptionsdata} dataset on MJHQ-30K~\cite{li2024playground}, which contains a higher proportion of highly-aesthetic images.

We also compare to the reported results from UniD3~\cite{hu2023unified}, which most closely resembles our work.\ifthenelse{\boolean{arxiv}}{}{ UniD3 released their inference code (\href{https://github.com/mhh0318/UniD3}{github.com/mhh0318/UniD3}) and a CUB200 checkpoint for unconditional generation, but did not release training code and we were unable to reproduce their results. We contacted the authors on this but we have not heard back. Therefore, we compare against their reported numbers on CUB200 and, as we show in \cref{tab:cub_fid}, our method obtains a significantly lower FID with only $51\%$ of the model parameters.

Aside from this, we note some differences between our works below:

\begin{enumerate}
    \item UniD3 uses a hybrid of a uniform and absorbing schedule, while we only use an absorbing schedule. We find the absorbing schedule to be significantly more training efficient than the uniform schedule. For instance we find that our 115M parameter model has an overall NLL of 6.1 at 60k steps with an absorbing schedule but 6.59 with a uniform schedule—with similar findings by several previous works including~\cite{Austin2021}.
    \item UniD3 decouples the modalities within the model, with a “mutual-attention” mechanism between modalities, and in loss computation. Instead, we simply use standard self-attention with a single weighted cross-entropy loss term ~\cref{eq:mdlm} and show that this scales to a 1.4B parameter model.
\end{enumerate}
}

\begin{table}[H]
    \centering
    \small
    \setlength{\tabcolsep}{3.5pt}
    \begin{tabular}{lccccccc}
        \toprule
        \textbf{Method} & \textbf{Sing. Obj.} & \textbf{Two Obj.} & \textbf{Counting} & \textbf{Colors} & \textbf{Position} & \textbf{Color Attr.} & \textbf{Overall} \\ \midrule
        SDv1.5~\cite{rombach2022high}   & 0.97 & 0.38 & 0.35 & 0.76 & 0.04 & 0.06 & 0.43 \\
        CoDI~\cite{tang2024any}   & 0.89 & 0.16 & 0.16 & 0.65 & 0.02 & 0.01 & 0.31 \\
        Lumina-mGPT~\cite{liu2024lumina}  & - & - & - & - & - & - & 0.32 \\
        \midrule
        \model{}  & 0.92 &  0.47 & 0.15 & 0.67 & 0.13 & 0.19 & 0.42 \\
        \bottomrule
    \end{tabular}
    \caption{We evaluate \model{} on the GenEval~\cite{ghosh2023genevalobjectfocusedframeworkevaluating} benchmark.}
    \label{tab:geneval}
\end{table}

\begin{table}[H]
    \centering
    \begin{tabular}{lcc}
        \toprule
        \textbf{Method} & \textbf{MSCOCO-30K FID} $\downarrow$ & \textbf{MJHQ-30K FID} $\downarrow$ \\
        \midrule
        SDv1.5~\cite{rombach2022high} & 11.12 & - \\
        CoDi~\cite{tang2024any} & \textbf{22.26} & 19.87 \\
        UniD3~\cite{hu2023unified}\protect\footnotemark & 25.11 & - \\
        \midrule
        \model{} (Ours) & 23.86 & \textbf{18.67} \\
        \bottomrule
    \end{tabular}
    \parbox{0.7\linewidth}{
        \centering
        \caption{We evaluate the 1.4B version of \model{} on FID. We use evaluate on MS-COCO 30K~\cite{chen2015microsoftcococaptionsdata} and MJHQ-30K~\cite{li2024playground}.}
        \label{tab:fid_scores_large}
    }
\end{table}

\begin{table}[H]
    \centering
    \begin{tabular}{lcc}
        \toprule
        \textbf{Method} & \textbf{Params} & \textbf{CUB200 FID} $\downarrow$ \\
        \midrule
        UniD3~\cite{hu2023unified} & 637M & 17.38 \\
        \midrule
        \model{} (Ours) & 330M & \textbf{11.03}\\
        \bottomrule
    \end{tabular}
    \parbox{0.7\linewidth}{
        \centering
        \caption{We compare our model to~\cite{hu2023unified} on CUB200.}
        \label{tab:cub_fid}
    }
\end{table}


\footnotetext{Trained only on MS-COCO. Other works listed in this table trained on a broader set of datasets (possibly including MS-COCO). In most cases, training on additional datasets likely harms dataset-specific FID.}

\newpage
\section{Understanding the effect of Classifier Free Guidance (CFG)}
\label{sec:cfg_study}
In Table \ref{tab:cond}, we observe that CFG is a significant factor in the performance difference between \model{} and the AR baseline. We hypothesize that this is because CFG is most useful in decoding the first few tokens, with diminishing utility in later tokens. To examine this, we look at intermediate predictions by storing $\arg\max \, p_\theta (x_0 \mid x_t)$ at each sampling step. As an AR model cannot directly capture this distribution without an intractable rollout, we opt to use the same \model{} model but with an autoregressive inference strategy, decoding from left to right. This allows us to directly compare the performance of different inference strategies and how they interact with classifier-free guidance.

\begin{figure}[H]
    \vspace{-5pt}
    \centering
    \begin{minipage}{0.48\textwidth}
        \centering
        \includegraphics[width=\linewidth]{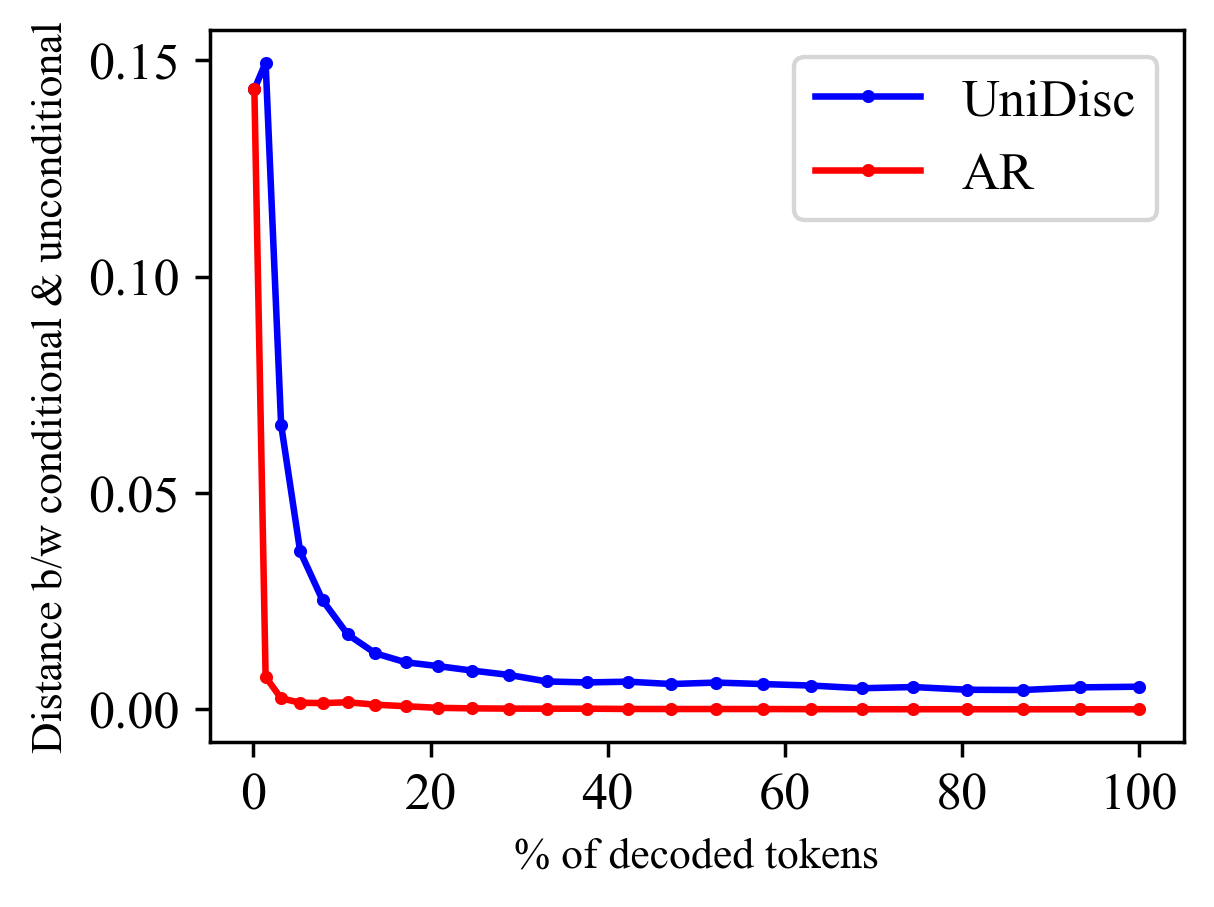}
        \caption{L2 distance between unconditional and conditional logits on currently masked tokens as sampling steps increase.}
        \label{fig:cfg_dist_vs_pct_tokens}
    \end{minipage}\hfill
    \begin{minipage}{0.48\textwidth}
        \centering
        \begin{tabular}{c|c}
        \hline
        Steps       & CLIP Score \\ \hline
        $\left[1-3 \right]$ & 0.301 \\
        $\left[12-14 \right]$ & 0.293 \\
        $\left[22-24 \right]$ & 0.283 \\
        \textbf{All (24)}     & 0.312 \\ \hline
        \end{tabular}
        \caption{Comparing CLIP scores by applying CFG only on specific steps. This shows CFG has the most impact on the initial denoising steps (total steps = 24).}
        \label{tab:cfg_select_tokens}
    \end{minipage}
    \vspace{-8pt}
\end{figure}

We visualize this in Figure \ref{fig:cfg_dist_vs_pct_tokens}, where we visualize the difference between the conditional and unconditional image generated at different percentages of decoded tokens. We notice two things: (a) the difference diminishes as more tokens are decoded and (b) \model{} consistently has higher distances between the logits than AR, which flattens out more quickly.

Intuitively, this means \model{} extracts much more discriminating signal from CFG compared to AR. We believe this is because \model{} has much more flexibility to decode tokens initially based on confidence, compared to AR which is forced to decode in a left to right manner and thus, can course correct quickly and more effectively. This can be seen in Table \ref{tab:cfg_select_tokens}, where we selectively apply CFG only on a few steps and notice that CLIP score when CFG is applied on steps 1-3 almost matches applying CFG on all, while applying on the last few steps doesn't affect things much at all.

Given the differences in CFG between \model{} and AR models, we conduct a hyperparameter sweep over guidance scales in Figure \cref{fig:cfg_weight_ablation}. We compute FID and CLIP scores over four datasets, and at both 115/340M parameters. We find that our AR baseline benefits from a weight of $w=0.5$ but sees far less improvement than \model{} with CFG. For \model{}, we choose an overall weight of $w=1.5$, but note that the CLIP score scales cleanly with the guidance scale, demonstrating the trade-off between visual quality and prompt adherence.

\begin{figure}[H]
    \vspace{-3pt}
    \centering
    \includegraphics[width=1.0\linewidth]{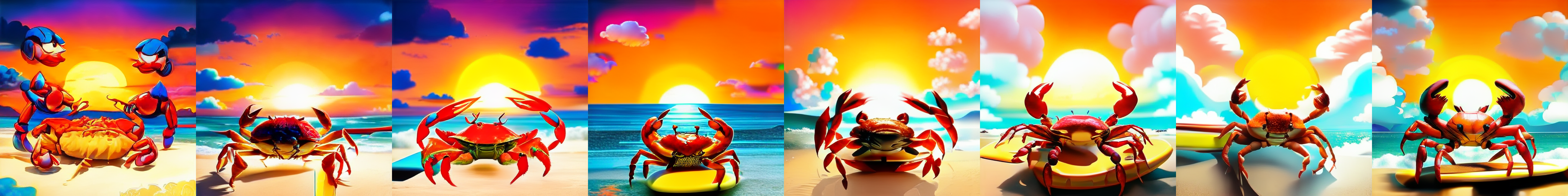}
    \caption{We show the effect of classifier-free guidance from left-to-right, starting with $w=0$, and increasing linearly to $w=8$ on the right, where output logits are $l_{\operatorname{cfg}}=(1+w)l_{\operatorname{cond}} + w*l_{\operatorname{uncond}}$.\\ Caption: "crab meditating, surfboard, orange sun setting, rainbow clouds, zen beach"}
    \label{fig:cfg_qualitative}
\end{figure}

Finally, in \cref{fig:cfg_qualitative}, we show the effect of CFG on the generated image. We increase the weight of the classifier-free guidance from $w=0$ to $w=8$ and observe the effect on the generated image.

\section{Inference: Generation time vs. batch size}
\begin{figure}[H]
    \centering
    \includegraphics[width=\linewidth]{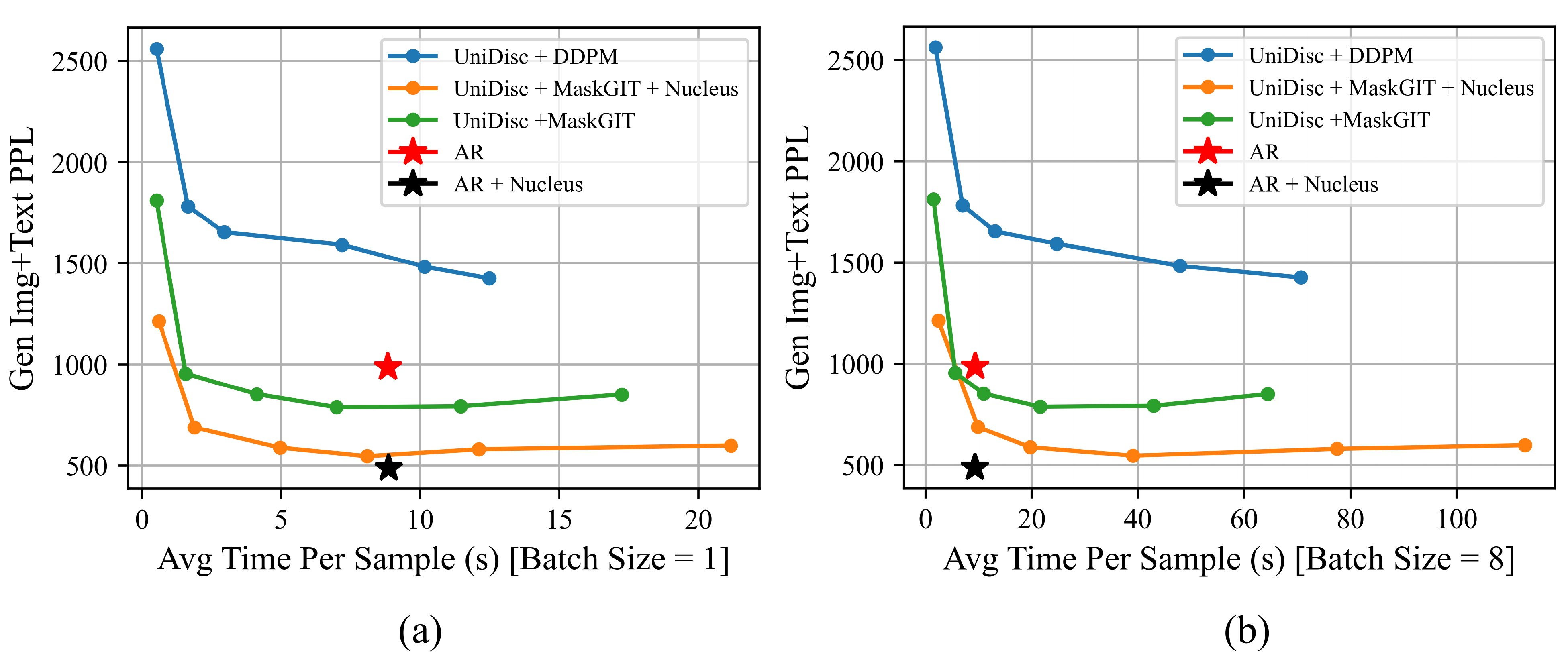}
    \caption{Generative Perplexity vs. Time with various models and sampling strategies.}
    \label{fig:batch_size_times}
\end{figure}

We analyze the quality of the generation versus time in Figure \ref{fig:batch_size_times}. We make a similar observation as in prior work~\cite{zivMaskedAudioGeneration2024,gatDiscreteFlowMatching2024a} on discrete diffusion, finding that the ability to obtain predictions with varying sampling steps allows lower latencies. However, with current implementations, KV caching in AR models results in higher throughput as the batch size increases. This tradeoff can be explained by looking at the number of function evaluations (NFEs) and the cost of each in both cases. In AR generation w/KV caching, we have a fixed NFE, but each forward pass is substantially less expensive than in the NAR case. In contrast, in NAR, we can use substantially fewer NFEs, but each is more costly. Modern GPUs only reach peak throughput at larger batch sizes~\cite{chittyvenkata2024llminferencebenchinferencebenchmarkinglarge}; as we decrease the batch size, the difference in computation per function evaluation diminishes, resulting in NAR having favorable performance.

\newpage


\end{document}